\documentclass[runningheads]{llncs}

\usepackage{eccv}

\usepackage{eccvabbrv}

\usepackage{graphicx}
\usepackage{booktabs}

\usepackage[accsupp]{axessibility}  

\usepackage{multirow}
\usepackage{adjustbox}
\usepackage{float}
\usepackage{wrapfig}

\usepackage{hyperref}

\usepackage{orcidlink}

\begin{document}

\title{Interpretation-Oriented Cloud Removal via Observation-Anchored Residual Flow with Geo-Contextual Alignment} 

\titlerunning{Interpretation-Oriented Cloud Removal}

\author{
Ziyao~Wang\inst{1,*}\orcidlink{0009-0004-7467-737X} \and
Maonan~Wang\inst{1,2}\orcidlink{0000-0001-5407-0416} \and
Yucheng~He\inst{1}\orcidlink{0009-0001-5730-9688} \and
Xianping~Ma\inst{3,\dagger}\orcidlink{0000-0002-2180-2964} \and
Ziyi~Wang\inst{1}\orcidlink{0009-0009-7058-0219} \and
Hongyang~Zhang\inst{1}\orcidlink{0009-0007-2617-3638} \and
Yirong~Cheng\inst{2}\orcidlink{0000-0001-8325-3596} \and
Man-on~Pun\inst{1,\dagger}\orcidlink{0000-0003-3316-5381}
}

\authorrunning{Z.~Wang et al.}

\institute{
School of Science and Engineering, The Chinese University of Hong Kong, Shenzhen, Shenzhen, China \and
Shanghai Ai Lab, Shanghai, China \and
Faculty of Geosciences and Engineering, Southwest Jiaotong University, Chengdu, China
}

\maketitle
\begingroup
\renewcommand{\thefootnote}{}
\footnotetext{$*$ First author. $\dagger$ Co-corresponding authors.}
\endgroup

\begin{abstract}
Cloud removal (CR) is essential for optical remote sensing, serving as a prerequisite for reliable downstream interpretation, such as semantic segmentation and change detection. However, existing CR approaches often prioritize visual realism while overlooking their impact on subsequent analytical tasks, leading to semantic drift and degraded downstream performance. To address this issue, we propose Geo-Anchored Cloud Removal (GACR), a unified framework that jointly ensures faithful reconstruction and robust interpretability. At its core, GACR incorporates Observation-Anchored Residual Flow (OAR-Flow), which reformulates CR as a physically grounded residual inversion process. By anchoring the generative trajectory to the cloudy observation rather than pure noise, OAR-Flow enables fast, stable, and faithful reconstruction. To further preserve semantic structures critical for downstream interpretation, GACR integrates Geo-Contextual Prior Alignment (GCPA) to constrain the reconstruction within a semantic manifold induced by a Vision Foundation Model (VFM). Consequently, GACR strictly maintains the spatial-semantic integrity of complex landscapes. Extensive experiments across six CR datasets and twelve downstream tasks demonstrate that GACR produces superior reconstruction quality while consistently improving downstream task accuracy. The code is available at \url{https://github.com/wzy6055/GACR}.

\end{abstract}

\section{Introduction}
\label{sec:intro}

Optical satellite imagery constitutes a primary data source for a broad spectrum of Earth observation applications, including urban development monitoring, resource management, and land-cover mapping \cite{zhu2017deep, astruc2024omnisat, guo2024skysense, zhu2025skysense}. The reliability of these applications fundamentally depends on accurate surface representation and semantic consistency in the observed imagery. However, the presence of clouds severely limits the usability of optical imagery by obscuring surface information and introducing uncertainty in analyses \cite{li2022cloud, king2013spatial}. As a result, cloud removal (CR) has evolved from a simple preprocessing operation to a critical component in the remote sensing interpretation pipeline, motivating extensive efforts toward generating cloud-free and semantically reliable imagery \cite{ebel2022sen12ms, king2013spatial}.

Recent advances in deep learning-based CR methods can be broadly categorized into denoising-based \cite{mehri2021mprnet, zamir2022restormer, zhou2024adapt} and generative-based approaches \cite{ma2023cloud, liu2025effective, sui2024diffusion, zou2024diffcr}. Denoising-based methods typically treat cloud occlusion as additive residual noise and learn a direct mapping to the underlying clear image. However, these methods are built upon the assumption that the residual noise follows a Gaussian prior distribution \cite{wang2025downstream}. Under heavy occlusion, where surface information is largely obscured rather than merely perturbed, such assumptions often lead to structural ambiguity and over-smoothed reconstructions. In contrast, generative approaches, particularly diffusion-based models, demonstrate a strong capability in synthesizing visually plausible textures under severe cloud coverage. Yet, their inherent stochastic sampling process lacks explicit observation anchoring, frequently resulting in geographically inconsistent structures or semantic drift in heavily occluded regions. While visually realistic, such hallucinated details may contradict the true land-cover distribution, thereby undermining reliability in downstream interpretation tasks.

Beyond the lack of observation anchoring in generative modeling, a more fundamental challenge lies in the visual-fidelity-oriented optimization paradigm underlying most existing CR methods. Current approaches primarily minimize pixel-level discrepancies against cloud-free references, implicitly equating visual fidelity with semantic correctness. Such objectives encourage aggressive artifact removal to maximize conventional image quality metrics (e.g., PSNR, SSIM) \cite{chen2025unirestore, wang2025downstream}, yet provide no explicit constraint on preserving task-relevant structural and categorical information. As a result, reconstructed regions may appear visually plausible while deviating from the true geographical context, leading to subtle but critical semantic inconsistencies. When deployed in downstream applications, such as semantic segmentation, these inconsistencies accumulate and translate into degraded representational reliability. This misalignment between low-level restoration objectives and high-level interpretative requirements limits the practical utility of current CR methods.

To address these challenges, we propose Geo-Anchored Cloud Removal (GACR), a unified framework built upon Observation-Anchored Residual Flow (OAR-Flow). Instead of initiating generation from pure noise, OAR-Flow starts from the cloudy observation and models CR as a physically grounded residual inversion process. By initializing the generative trajectory with structured perturbations around the observed image, the model adapts its behavior according to cloud opacity. In thin-cloud regions, where surface signals remain partially observable, the observation dominates the dynamics, guiding the model to perform physical inversion and preserve subtle yet authentic surface cues. In contrast, under thick cloud coverage where information is largely obscured, residual perturbations provide generative flexibility, enabling controlled semantic completion. This observation-anchored mechanism shortens the generative trajectory and suppresses unnecessary stochastic exploration, ensuring that reconstructed structures remain spatially aligned with the original observation while avoiding geographically implausible artifacts.

\begin{figure}[!t]
	\centering
	\begin{subfigure}[t]{0.26\linewidth}  
		\centering
		\includegraphics[width=\linewidth]{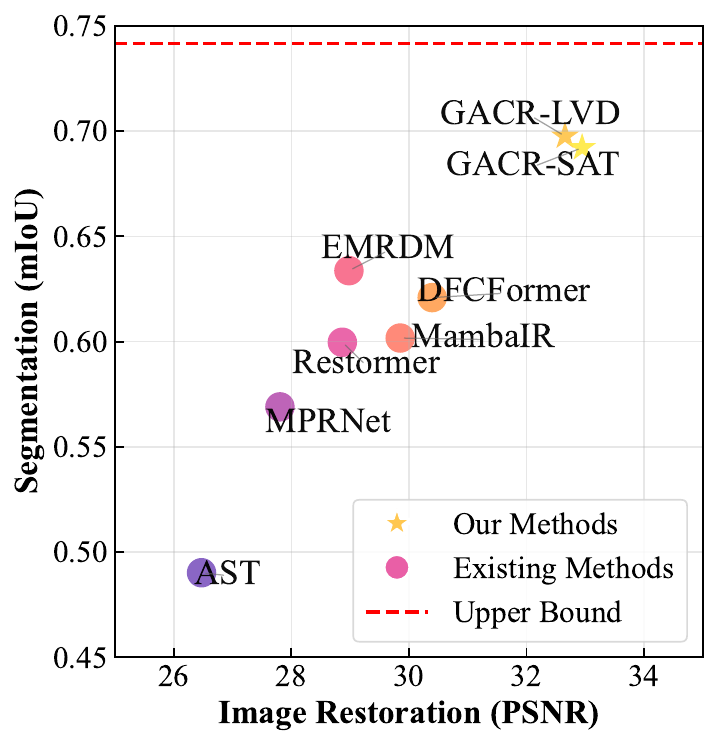}
		\caption{PSNR and mIoU comparison.}
		\label{fig:psnr_vs_miou}
	\end{subfigure}
	\begin{subfigure}[t]{0.26\linewidth}  
		\centering
		\includegraphics[width=\linewidth]{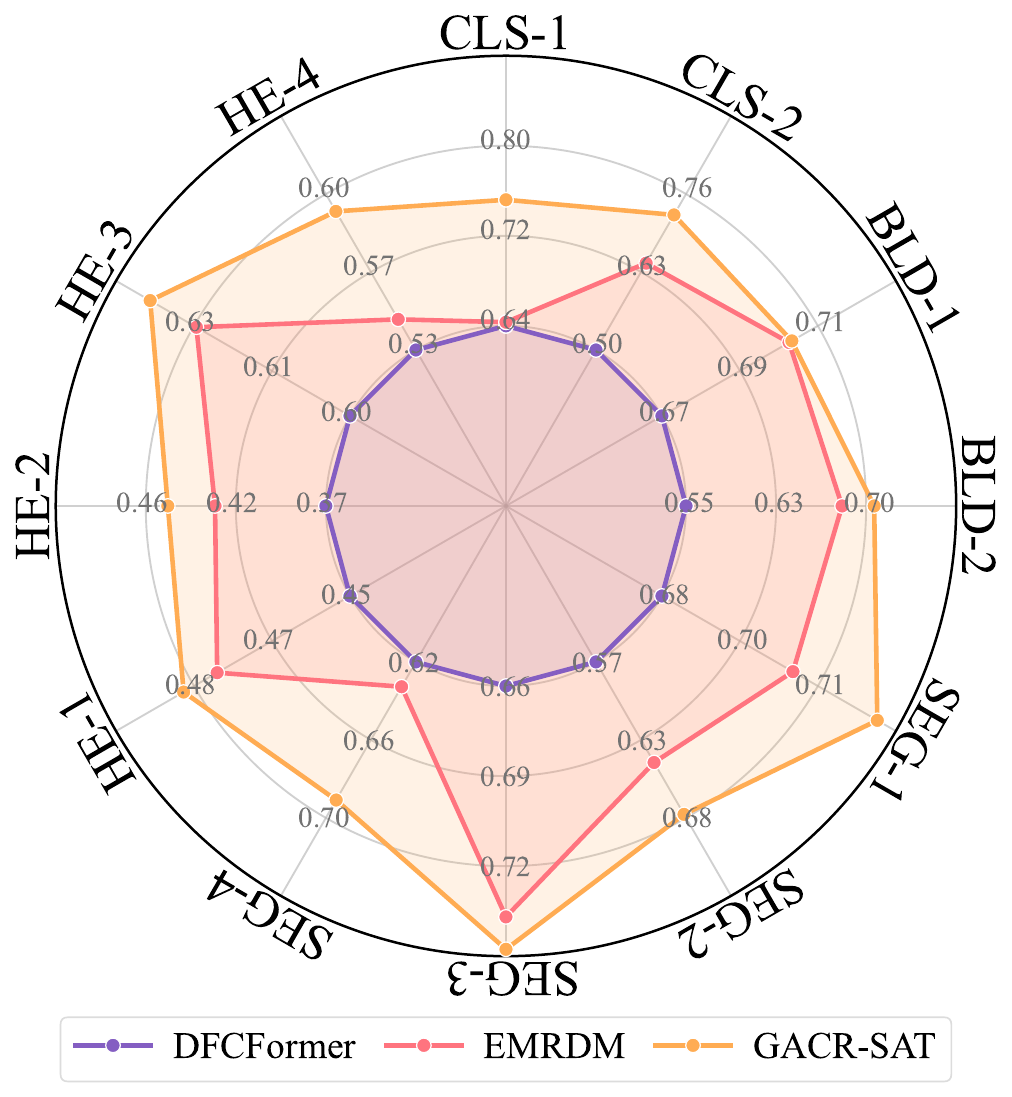}
		\caption{Downstream performance.}
		\label{fig:radar}
	\end{subfigure}
    	\begin{subfigure}[t]{0.42\linewidth} 
		\centering
        \includegraphics[width=\linewidth]{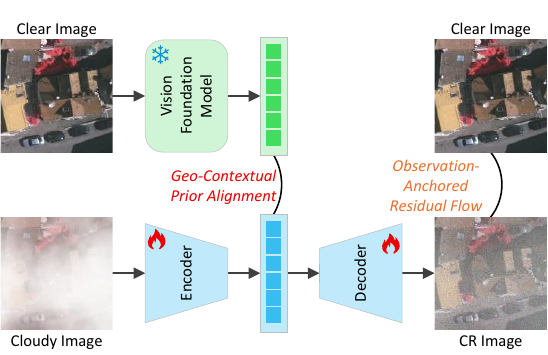}
		\caption{OAR-Flow and GCPA.}
		\label{fig:highlight3}
	\end{subfigure}
	\caption{(a) Comparison with existing methods in terms of PSNR and mIoU on Vaihingen-CR-Thick. (b) Performance across 12 downstream tasks, where the outermost ring denotes the upper bound. (c) GACR reconstructs cloud-free imagery from cloudy observations via OAR-Flow, while GCPA constrains the generative process within a geo-contextually consistent semantic manifold.}
	\label{fig:highlight}
\end{figure}

While physical anchoring stabilizes the generative dynamics, visual fidelity alone remains insufficient to guarantee semantic reliability for downstream interpretation. To bridge the gap between low-level restoration and high-level semantic preservation, we introduce Geo-Contextual Prior Alignment (GCPA), which leverages representational priors from a pretrained Vision Foundation Model (VFM) to guide cloud removal. Rather than optimizing solely for pixel-level similarity, GCPA aligns dense representations within a VFM-induced semantic manifold, encouraging reconstructed regions to remain coherent with their surrounding geographical context. This alignment is enforced through the proposed Geo-Contextual Integrity Loss (GCI Loss), which explicitly regularizes the generative process to preserve task-relevant structural patterns and category-specific semantic signatures. By jointly integrating OAR-Flow and GCPA, our framework produces cloud-free reconstructions that maintain both visual fidelity and semantic integrity. Extensive experiments across six CR datasets and twelve downstream tasks demonstrate that GACR consistently improves both reconstruction quality and downstream interpretation accuracy, with PSNR gains reaching 3.3 dB, semantic segmentation improvements of 3.1 mIoU, and approximately 5× faster convergence, effectively mitigating semantic distortion, as shown in \cref{fig:highlight}.

The main contributions of this work are:
\begin{itemize}
\item We formulate cloud removal as an interpretation-oriented generative inversion problem and propose OAR-Flow, which grounds generation in the observed cloudy image to achieve stable and faithful reconstruction.

\item We introduce GCPA, which constrains reconstruction within a semantic manifold induced by a VFM to preserve task-relevant structural patterns and category-specific information.

\item Extensive experiments across six benchmarks and twelve downstream tasks demonstrate that GACR consistently improves reconstruction quality while yielding higher accuracy in downstream interpretation tasks.
\end{itemize}

\section{Related Work}
\label{sec:rel}
\subsection{Cloud Removal Method}

CR is a fundamental preprocessing step in Earth observation, aiming to recover cloud-free imagery for reliable surface analysis. With deep learning, data-driven CR methods have greatly improved visual quality and adaptability, and can be broadly grouped into residual-prediction-based and generation-based methods. The former estimates cloud residuals or clean reflectance using CNNs \cite{li2019thick, zi2021thin}, Transformers \cite{jin2024rfe, wang2025downstream}, or Mamba architectures \cite{liu2025cr, pan2025m, guo2025mambair, gu2025acl}. However, they commonly rely on the cloud-as-residual assumption, which oversimplifies complex atmospheric scattering and limits recovery in regions severely occluded by thick clouds. Generation-based models instead synthesize cloud-free images through adversarial or diffusion mechanisms. Early GAN-based methods \cite{bermudez2018sar, enomoto2017filmy, grohnfeldt2018conditional, ma2023cloud, singh2018cloud} improved perceptual realism but often suffered from instability and artifacts, whereas recent diffusion-based approaches \cite{zou2024diffcr, sui2024diffusion, liu2025effective, silva2025cloud} achieve better fidelity and convergence. Nevertheless, most CR models are still mainly optimized for pixel-level similarity, offering limited semantic constraints for downstream interpretation tasks \cite{chen2025unirestore, wang2025downstream}. In contrast, our framework introduces OAR-Flow and GCPA to jointly promote physically grounded reconstruction and semantic integrity.

\subsection{Diffusion and Flow Model}

Diffusion models have recently shown remarkable success in image generation \cite{wang2025reconciling, jeong2025latent, yang2025fam, feng2025gps}, evolving from DDPM \cite{ho2020denoising, nichol2021improved} and DDIM \cite{song2020denoising} to score-based SDE/ODE formulations \cite{song2021scorebased, karras2022elucidating}. Through iterative denoising of Gaussian noise, they achieve higher fidelity and diversity than GAN-based \cite{goodfellow2020generative} and VAE-based \cite{kingma2013auto} paradigms. Early studies mainly used U-Net architectures \cite{zou2024diffcr, sui2024diffusion}, while recent Diffusion Transformers (DiT) \cite{peebles2023scalable} improve scalability. However, many transformer-based diffusion models work in latent space \cite{rombach2022high, peebles2023scalable, ma2024sit}, which limits pixel-level restoration. HDiT \cite{crowson2024scalable} therefore enables direct pixel-space reconstruction, with MRDM \cite{luo2023image} and EMRDM \cite{liu2025effective} further adapting diffusion to CR scenarios. Nevertheless, diffusion models often require long stochastic sampling trajectories and high computational cost \cite{ho2020denoising}. Flow matching methods \cite{lipman2022, albergo2023stochastic, albergo2023building, liu2022} reformulate diffusion as deterministic probability flow ODEs for more efficient trajectory learning. Different from existing flow-based CR methods, OAR-Flow introduces an observation-conditioned residual formulation that grounds generative dynamics in the cloudy input rather than unconditional noise.

\begin{figure*}[tbp]
	\centering
	{\includegraphics[width=\linewidth]{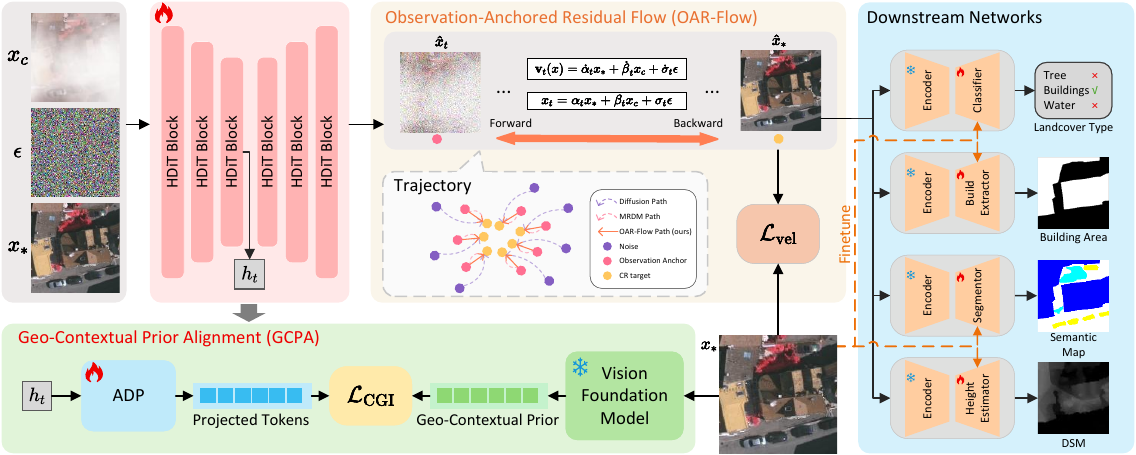}}
	\caption{Overview of the proposed GACR framework. (1) OAR-Flow reconstructs cloud-free imagery from cloudy observations via an observation-anchored residual trajectory, replacing pure noise initialization with a physically grounded anchor and enabling stable deterministic flow dynamics supervised by $\mathcal{L}_{vel}$. (2) GCPA leverages a pretrained Vision Foundation Model to extract geo-contextual representations from clean images, enforcing semantic consistency through the $\mathcal{L}_{GCI}$ and constraining the generative process within a coherent feature manifold. (3) Downstream networks trained on cloud-free data are used to evaluate the semantic fidelity and interpretation reliability of the reconstructed CR outputs.}
	\label{fig:network_sp1}
\end{figure*}

\section{Methodology}
\label{sec:met}

This section presents the proposed GACR framework illustrated in \cref{fig:network_sp1}. We start by reviewing the theoretical background in \cref{subsec:pre}. We then introduce OAR-Flow in \cref{subsec:oar}, followed by GCPA in \cref{subsec:gcpa}. Finally, we describe the downstream evaluation protocols in \cref{subsec:down}.

\subsection{Preliminaries}
\label{subsec:pre}

A diffusion process with an initial distribution $p_0(x)$ can be described by the following stochastic differential equation (SDE) \cite{song2021scorebased, lu2022dpm, de2022riemannian} for $t \in [0, T]$:
\begin{equation}
    dx = f(x, t)\,dt + g(t)\,d\mathbf{w}, \quad x(0) \sim p_0(x),
\end{equation}
where $f(\cdot,\cdot)$ and $g(\cdot)$ denote the drift and diffusion coefficients, and $\mathbf{w}$ is a standard Wiener process. The corresponding reverse-time SDE is given by:
\begin{equation}
    dx = \big[f(x, t) - g(t)^2 \nabla_x \log p_t(x)\big]dt + g(t)\,d\hat{\mathbf{w}},
\end{equation}
where $\hat{\mathbf{w}}$ is the reverse-time Wiener process and $\nabla_x \log p_t(x)$ denotes the score function of the marginal distribution $p_t(x)$. In image restoration problems, the objective is to model the conditional distribution between degraded and clean images by constructing a continuous trajectory that links the two states.

To better characterize structured degradation, several works introduced modified diffusion dynamics that interpolate between a reference state and the target distribution via mean-reverting formulations \cite{luo2023image, liu2025effective}. A representative forward process can be written as
\begin{equation}
    dx = \theta_t (\mu - x)\,dt + \sigma_t\,d\mathbf{w},
\end{equation}
where $\mu$ denotes a reference state and $\theta_t, \sigma_t$ control the drift strength and stochastic noise intensity, respectively. This formulation provides an interpretable trajectory that gradually transports samples toward a designated state, offering a structured alternative to purely noise-driven diffusion. The corresponding reverse process can be expressed in either SDE or ODE form depending on the parameterization \cite{luo2023image, liu2025effective}. Such formulations provide a flexible basis for designing conditional generative dynamics, which we further specialize for CR in the following section.

\subsection{Observation-Anchored Residual Flow}
\label{subsec:oar}

\noindent\textbf{Forward Process.}
Unlike unconditional generation, CR aims to recover the clean surface image $x_*$ from a structured degradation $x_c$ (cloudy observation). In remote sensing imagery, cloud coverage is not random noise but a spatially varying atmospheric scattering layer that partially or fully obscures surface reflectance. To explicitly model this conditional relationship, OAR-Flow constructs a continuous trajectory that transports samples between the clean state $x_*$ and the cloudy observation $x_c$ under structured perturbation, as illustrated in \cref{fig:network_sp1}.

We define the forward interpolant as:
\begin{equation}
    x_t = \alpha_t x_* + \beta_t x_c + \sigma_t \epsilon,
    \label{eq:oar_forward}
\end{equation}
where $\epsilon \sim \mathcal{N}(0, I)$ denotes Gaussian noise, and $\alpha_t$, $\beta_t$, and $\sigma_t$ are time-dependent coefficients satisfying:
\begin{equation}
\begin{aligned}
\alpha_0 &= 1, & \beta_0 &= 0, & \sigma_0 &= 0,\\
\alpha_T &= 0, & \beta_T &> 0. &&
\end{aligned}
\end{equation}
Here, $x_c$ acts as an observation anchor. In thin-cloud regions, where surface signals remain partially observable, the contribution of $x_c$ preserves low-frequency and structural cues. In thick-cloud regions, the stochastic component $\epsilon$ provides generative flexibility for semantic completion. This formulation explicitly reflects the opacity-aware characteristics of cloud degradation.

In practice, we adopt a simple linear schedule:
\begin{equation}
    \alpha_t = 1 - t, \quad
    \beta_t = \rho t, \quad
    \sigma_t = t,
\end{equation}
where $\rho$ controls the strength of the observation anchor relative to stochastic perturbations, ensuring that the trajectory remains grounded in the cloudy observation while retaining flexibility in severely occluded regions.

\noindent\textbf{Backward Process.}
The marginal distribution $p_t(x)$ induced by \cref{eq:oar_forward} satisfies the transport equation \cite{albergo2023stochastic, ma2024sit}:
\begin{equation}
    \partial_t p_t(x) + \nabla_x \cdot \big( \mathbf{v}_t(x) p_t(x) \big) = 0,
    \label{eq:oar_transport}
\end{equation}
where $\mathbf{v}_t(x)$ denotes the velocity field of the deterministic probability flow.

Following the flow matching formulation, the ideal velocity field can be expressed as:
\begin{equation}
    v_t(x)=\mathbb{E}[\dot{\alpha}_t x_*+\dot{\beta}_t x_c+\dot{\sigma}_t\epsilon \mid x_t=x],
    \label{eq:oar_velocity}
\end{equation}
which explicitly decomposes the dynamics into three components: clean target guidance, observation anchoring, and stochastic perturbation. This residual decomposition differs from purely noise-driven flows by incorporating structured observational information into the trajectory. The correspondence between the marginal distribution $p_t(x)$ and the velocity formulation in \cref{eq:oar_velocity} follows from the transport equation in \cref{eq:oar_transport}, and the detailed derivation is provided in Appendix~A.

We parameterize the velocity field using a neural network $\mathbf{u}_t$:
\begin{equation}
    \mathbf{u}_t(x) = \mathrm{Net}_{\theta}(x_t, t, x_c),
\end{equation}
where $x_c$ is provided as a condition to preserve spatial alignment with the observed cloudy image.

During inference, the clean estimate is obtained by integrating the deterministic flow from $t=T$ to $t=0$:
\begin{equation}
    \hat{x}_{t-1} = \hat{x}_t + \Delta t \, \mathbf{u}_t(\hat{x}_t),
\end{equation}
yielding the final reconstruction $\hat{x}_* = \hat{x}_0$. Because the trajectory remains anchored to $x_c$ throughout integration, the reconstructed surface structures are spatially consistent with the original observation, effectively reducing geographically implausible artifacts.

\subsection{Geo-Contextual Prior Alignment}
\label{subsec:gcpa}

To jointly ensure reconstruction fidelity and semantic integrity in CR, we optimize OAR-Flow under two complementary objectives: a velocity matching loss $\mathcal{L}_{\mathrm{vel}}$ in pixel space and a geo-contextual consistency constraint loss $\mathcal{L}_{\mathrm{GCI}}$ in the representation space.

\noindent\textbf{Velocity Matching Loss.}
The primary supervision for OAR-Flow is to match the analytical velocity defined in \cref{eq:oar_velocity}. The network $\mathbf{u}_t$ is trained to approximate the target velocity field:
\begin{equation}
\begin{aligned}
\mathcal{L}_{\mathrm{vel}}
&= \mathbb{E}\!\left[\!\left\|\mathbf{u}_{t}(x_t)-\mathbf{v}_t(x_t)\right\|^{2}\!\right] \\
&= \mathbb{E}_{x_{*},\epsilon,x_c}\!\left[\!\left\|
\mathbf{u}_{t}(x_t)-\dot{\alpha}_{t}x_{*}-\dot{\beta}_{t}x_c-\dot{\sigma}_{t}\epsilon
\right\|^{2}\!\right].
\end{aligned}
\end{equation}

This objective ensures that the learned flow follows the observation-anchored residual trajectory defined in OAR-Flow. The network is implemented using an HDiT-based backbone \cite{crowson2024scalable}, receiving the current state $x_t$, timestep $t$, and the cloudy observation $x_c$ as condition, and predicting the deterministic velocity toward the clean state.

\noindent\textbf{Geo-Contextual Integrity Loss.}
While velocity supervision guarantees pixel-level reconstruction consistency, it does not explicitly enforce preservation of land-cover semantics. In remote sensing imagery, geographical context exhibits strong structural continuity (e.g., forests form contiguous regions and urban layouts follow spatial regularity). To maintain such geo-contextual coherence, we introduce GCPA, implemented through the proposed GCI Loss.

Let $f_{\mathrm{vfm}}(\cdot)$ denote a pretrained VFM encoder. The geo-contextual prior of the clear image is defined as
\begin{equation}
    z_* = f_{\mathrm{vfm}}(x_*) \in \mathbb{R}^{B \times L \times D}.
\end{equation}

We extract the intermediate feature $h_t \in \mathbb{R}^{B \times C \times H \times W}$ from the bottleneck of the HDiT backbone and map it into the same representation space via an Adaptive Projector (ADP):
\begin{equation}
\begin{aligned}
    z_t &= \mathrm{ADP}(h_t) \\
        &= \mathrm{MLP}\big(\mathrm{RE}(\mathrm{AP}(h_t))\big),
\end{aligned}
\end{equation}
where $\mathrm{RE}(\cdot)$ and $\mathrm{AP}(\cdot)$ denote rearrangement and adaptive pooling operations, respectively.

The GCI Loss is defined as a patch-wise cosine similarity:
\begin{equation}
\mathcal{L}_{\mathrm{GCI}}
= - \mathbb{E}\!\left[\frac{1}{N} \sum_{n=1}^{N}
\frac{\langle z_*^{[n]}, z_t^{[n]} \rangle}
{\|z_*^{[n]}\|_2 \, \|z_t^{[n]}\|_2}
\right],
\end{equation}
where $n$ indexes spatial tokens. By aligning reconstructed features with the VFM-induced semantic manifold, this loss constrains the generative process to remain consistent with large-scale geographical structures, thereby reducing semantic drift in heavily occluded regions.

\noindent\textbf{Unified Objective.}
The overall training objective is:
\begin{equation}
    \mathcal{L} = \mathcal{L}_{\mathrm{vel}} + \lambda \mathcal{L}_{\mathrm{GCI}},
\end{equation}
where $\lambda$ balances reconstruction fidelity and geo-contextual integrity. Ablation analysis of the objective components is provided in Section~\ref{subsec:abla}.

\subsection{Downstream Networks}
\label{subsec:down}

We evaluate GACR on four remote sensing tasks: land-cover classification, building extraction, semantic segmentation, and height estimation as shown in Fig.~\ref{fig:network_sp1}. These tasks assess whether reconstructed CR results preserve task-relevant semantic and structural information.

We adopt a pretrained DINOv3 backbone \cite{simeoni2025dinov3} as a unified visual encoder for downstream evaluation. For each task, the backbone is frozen and paired with a lightweight decoder trained on clear images to establish reliable reference performance. Notably, the VFM used for GCPA and the downstream encoders are independent, ensuring that downstream evaluation does not share parameters or supervision signals with the upstream semantic guidance. The architectural details of the downstream networks are presented in Appendix~B.

\section{Experiments and Discussion}
\label{sec:exp}

\subsection{Implementation Details}
\label{subsec:impl}

\noindent{\textbf{Dataset.}}
To enable a unified evaluation of both reconstruction fidelity and downstream performance, we construct a comprehensive benchmark for joint assessment. Specifically, we employ six CR datasets, including two publicly available datasets: CUHKCR-EXT-GZ and CUHKCR-EXT-CS \cite{wang2025downstream}, along with four synthetic cloud datasets constructed by ourselves: Potsdam-CR-thin, Potsdam-CR-thick, Vaihingen-CR-thin, and Vaihingen-CR-thick. These datasets cover both real and simulated cloud scenarios, encompassing thin and thick cloud conditions to reflect varying levels of opacity and structural occlusion.

\noindent{\textbf{Downstream Tasks.}}
CUHKCR-EXT-GZ and CUHKCR-EXT-CS support land-cover classification (CLS-1 and CLS-2) and building extraction (BLD-1 and BLD-2), while the four synthetic datasets include semantic segmentation (SEG-1 to SEG-4) and height estimation (HE-1 to HE-4). These tasks evaluate whether reconstructed CR outputs preserve task-relevant semantic structures and spatial consistency. Further details on dataset construction and the cloud synthesis process are provided in Appendix~C.

\noindent{\textbf{Model Settings.}}  
For GCPA, we employ two pretrained DINOv3 variants as the VFM module: ViT-L/16-SAT-300M and ViT-L/16-LVD-1689M. Model configurations are denoted as GACR-SAT/p or GACR-LVD/p, where “SAT” and “LVD” indicate the corresponding VFM pretraining variants, and “/p” specifies the patch size used in OAR-Flow. In our experiments, the upstream VFM for geo-contextual guidance and the downstream encoder for evaluation are initialized from different pretrained weights, ensuring no parameter sharing between guidance and evaluation stages. This separation avoids potential bias arising from shared representation priors and enables a fair assessment of downstream improvements. To further verify robustness with respect to backbone selection, we additionally conduct experiments using heterogeneous downstream encoders, with detailed results reported in Appendix~D.

\noindent{\textbf{Evaluation Metrics.}}
For CR evaluation, we adopt PSNR and SSIM to measure reconstruction fidelity from complementary perspectives of distortion, perceptual similarity, and error magnitude. For downstream evaluation, we report accuracy (Acc.) for CLS, IoU for BLD, mIoU for SEG, and RMSE for HE. More implementation details of metrics and training settings are provided in Appendix~D.

\begin{table}[t]
\centering
\caption{Quantitative comparison across six CR datasets. The best and second-best scores are marked in \textbf{bold} and \underline{underline}, respectively.}
\begin{adjustbox}{width=1.0\textwidth, center}
\begin{tabular}{l  |c c |c c |c c |c c |c c |c c}
\toprule
\multirow{2}{*}{Model} & 
\multicolumn{2}{c|}{CUHKCR-EXT-GZ}& 
\multicolumn{2}{c|}{CUHKCR-EXT-CS} & \multicolumn{2}{c|}{Potsdam-CR-thin}& \multicolumn{2}{c|}{Potsdam-CR-thick}& \multicolumn{2}{c|}{Vaihingen-CR-thin}& \multicolumn{2}{c}{Vaihingen-CR-thick
}\\ 
 & PSNR$\uparrow$ & SSIM$\uparrow$ & 
 PSNR$\uparrow$ & SSIM$\uparrow$  & PSNR$\uparrow$ & SSIM$\uparrow$ & PSNR$\uparrow$ & SSIM$\uparrow$ & PSNR$\uparrow$ & SSIM$\uparrow$ & PSNR$\uparrow$ &SSIM$\uparrow$ \\ 
\midrule
MPRNet \cite{mehri2021mprnet} & 23.454 & 0.712 & 23.365 & 0.695  & 28.349 & 0.960 & 26.418 & 0.902 & 31.209 & 0.978 & 27.805 &0.935 
\\
Restormer \cite{zamir2022restormer} & 25.839 & 0.743 & 23.632 & 0.710  & 31.413 & 0.970 & 28.831 & 0.922 & 31.210 & 0.970 & 28.867 &0.922 
\\
AST \cite{zhou2024adapt} & 25.482 & 0.735 & 23.365 & 0.695  & 28.924 & 0.957 & 25.886 & 0.890 & 29.877 & 0.973 & 26.471 &0.916 
\\
MambaIR \cite{guo2025mambair} & 25.626 & 0.733 & 23.445 & 0.704  & 30.881 & 0.967 & 27.027 & 0.903 & 32.764 & 0.983 & 29.852 &0.945 
\\
DFCFormer \cite{wang2025downstream} & 25.816 & \underline{0.746}& 23.876 & 0.711  & 30.836 & 0.969 & 28.196 & 0.917 & 33.342 & 0.985 & 30.396 &0.951 
\\
EMRDM \cite{liu2025effective} & 25.862 & \textbf{0.747}& 23.736 & 0.712  & 30.335 & 0.972 & 27.199 & 0.923 & 33.620 & 0.988 & 28.979 &0.951 
\\
\midrule
GACR-SAT/2& 25.964 & 0.736 & 24.230 & 0.709  & 33.141 & 0.975& 30.578 & 0.934 & 36.293 & 0.990 & 33.018 &0.964 \\
GACR-LVD/2& 25.899 & 0.735 & 24.220 & 0.707  & 32.880 & 0.975& 30.308 & 0.932 & 35.749 & 0.989 & 32.656 &0.963 
\\
GACR-SAT/1& \textbf{26.100}& 0.744 & \textbf{24.354}& \underline{0.713} & \textbf{33.642}& \textbf{0.976}& \textbf{31.049}& \textbf{0.938}& \textbf{36.918}& \textbf{0.991}& \textbf{34.048}&\textbf{0.970}\\
GACR-LVD/1& \underline{26.059}& 0.744 & \underline{24.331}& \textbf{0.714} & \underline{33.305}& \underline{0.975}& \underline{30.586}& \underline{0.935}& \underline{36.590}& \underline{0.991}& \underline{33.808}&\underline{0.969}\\
\bottomrule
\end{tabular}
\end{adjustbox}
\label{tab:quantative_results}
\end{table}

\begin{figure*}[htbp]
	\centering
	{\includegraphics[width=\linewidth]{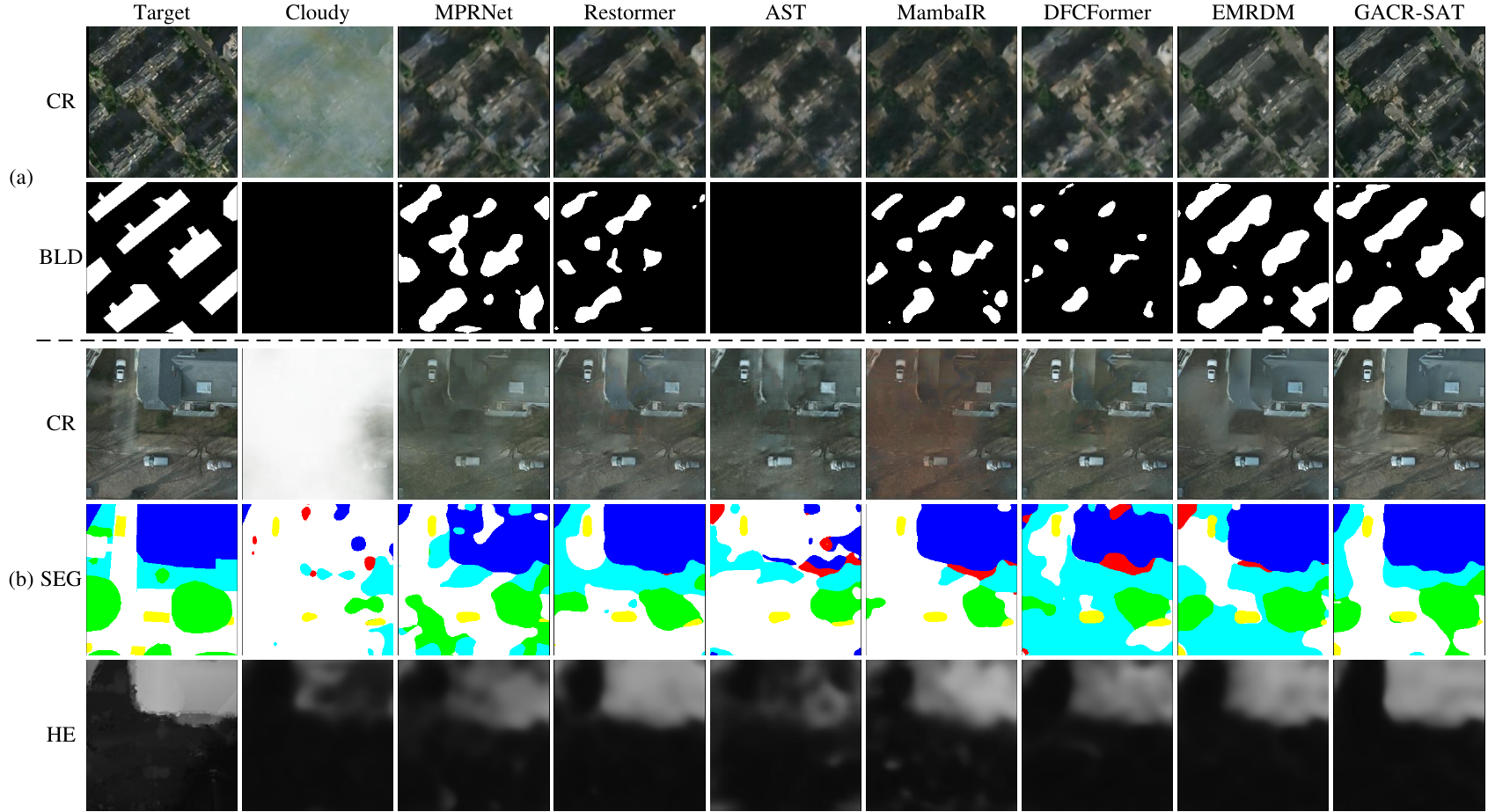}}
	\caption{Visualization of CR and downstream results. (a) The CR results on CUHKCR-EXT-GZ and the corresponding BLD results. (b) The CR results on Potsdam-CR-thick and the corresponding SEG and HE results.}
	\label{fig:viz_compare}
\end{figure*}

\subsection{CR Evaluation}
\label{subsec:crev}

This section evaluates the reconstruction fidelity of CR results. Quantitative comparisons are reported in \cref{tab:quantative_results}. On the CUHKCR-EXT datasets, GACR achieves highly competitive performance across most metrics. Except for a slightly lower SSIM on CUHKCR-EXT-GZ, GACR consistently surpasses existing approaches in terms of PSNR and RMSE, indicating a favorable balance between pixel-level accuracy and perceptual consistency. For example, on CUHKCR-EXT-CS, GACR-SAT/1 achieves a PSNR of 24.354, improving upon the strongest baseline by approximately 0.5 dB. Notably, the improvements are not limited to a single metric but are consistently reflected across multiple evaluation metrics, demonstrating the robustness of the proposed reconstruction strategy.

On the four synthetic datasets, GACR demonstrates clear advantages under both thin- and thick-cloud conditions. In thin-cloud scenarios, where surface structures remain partially observable, the observation-anchored formulation effectively preserves fine-grained textures and low-frequency spatial continuity, preventing unnecessary alterations to already reliable regions. This leads to the highest PSNR of 33.642 dB on Potsdam-CR-thin and 36.918 dB on Vaihingen-CR-thin with GACR-SAT/1. In thick-cloud settings, where large areas are severely occluded and lack direct visual cues, the residual generative component provides controlled semantic completion guided by the anchored trajectory. This mechanism enables the model to recover plausible yet geographically consistent structures, resulting in consistent improvements over prior methods across multiple quantitative metrics. The performance gains under heavy occlusion further validate the effectiveness of modeling CR as a physically grounded residual inversion process.

We further evaluate two patch-size configurations to examine the trade-off between reconstruction granularity and computational cost. Results indicate that GACR-SAT/2 already surpasses existing methods by a clear margin, confirming that the proposed framework remains effective even under coarser spatial partitioning. A smaller patch size (GACR-SAT/1) further enhances reconstruction fidelity by enabling finer spatial modeling and more detailed residual refinement. However, considering computational efficiency and fairness in comparison with existing baselines, we adopt $p=2$ in subsequent experiments unless otherwise specified. Qualitative comparisons are shown in \cref{fig:viz_compare}, where GACR produces sharper structural boundaries and fewer texture inconsistencies. Additional visualizations are provided in Appendix~E to further illustrate the stability of the reconstruction results across diverse cloud conditions.

\begin{table}[t]
\centering
\caption{Quantitative comparison of different tasks on downstream networks with ViT-L/16-LVD-1689M weight, including classification (CLS), building extraction (BLD), semantic segmentation (SEG), and height estimation (HE).
The best and second-best results are highlighted in \textbf{bold} and \underline{underline}, respectively.}
\begin{adjustbox}{width=1.0\textwidth, center}
\begin{tabular}{l  |c c  |c c  |c llc  |c cll}
\toprule
\multirow{2}{*}{Model} & 
CLS-1&CLS-2&BLD-1&BLD-2& 
SEG-1&  SEG-2&SEG-3&SEG-4&HE-1& HE-2& HE-3&HE-4\\ 
& Acc. $\uparrow$& Acc. $\uparrow$& IoU $\uparrow$& IoU $\uparrow$& 
 mIoU $\uparrow$&   mIoU $\uparrow$&mIoU $\uparrow$&mIoU $\uparrow$& RMSE $\downarrow$&  RMSE $\downarrow$& RMSE $\downarrow$&RMSE $\downarrow$\\ 
\midrule
Upper Bound& 0.882 & 0.916 & 0.718 & 0.762 & 0.733 &   0.733 &0.755 &0.755 & 1.868 &  1.868 & 1.477 &1.477 
\\
 Lower Bound& 0.746 & 0.739 & 0.669 & 0.596 & 0.657 & 0.490 & 0.677 & 0.550 & 2.144 & 2.703 & 1.672 &2.095 
\\
 End-to-End& 0.833 & 0.876 & 0.691 & 0.698 & 0.709 & 0.643 & 0.737 & 0.684 & 2.000 & 2.245 & 1.583 &1.830 
\\
\midrule
 MPRNet \cite{mehri2021mprnet}& 0.809 & 0.752 & 0.653 & 0.654 & 0.707 & 0.615 & 0.741 & 0.671 & 1.987 & 2.412 & 1.549 &1.757 
\\
Restormer \cite{zamir2022restormer} & 0.803 & 0.744 & 0.655 & 0.660 & 0.710 &   0.650 &0.718 &0.675 & 1.913 &  2.228 & 1.520 &1.681 
\\
AST \cite{zhou2024adapt} & \underline{0.813}& 0.767& 0.668 & 0.651 & 0.697 &   0.580 &0.736 &0.632 & 2.019 &  2.492 & 1.585 &1.885 
\\
MambaIR \cite{guo2025mambair} & 0.759 & \underline{0.778}& 0.660 & 0.648 & 0.713 &   0.620 &0.736 &0.671 & 1.948 &  2.350 & 1.552 &1.697 
\\
DFCFormer \cite{wang2025downstream} & 0.763 & 0.771 & 0.659 & 0.657 & 0.713 &   0.630 &0.722 &0.671 & 1.948 &  2.317 & 1.526 &1.687 
\\
EMRDM \cite{liu2025effective} & 0.776 & 0.726 & \underline{0.703}& 0.696 & 0.722 &   0.668 &0.747 &0.692& 1.902 &  2.133 & 1.512 &1.629 
\\
\midrule
GACR-SAT/2& \textbf{0.833}& \textbf{0.781}& \textbf{0.704}& \textbf{0.710}& \textbf{0.727}&   \textbf{0.699}&\textbf{0.750}&\textbf{0.737}& \underline{1.891}&  \textbf{2.014}& \textbf{1.482}&\textbf{1.554}\\
GACR-LVD/2& 0.806 & 0.768 & 0.702 & \underline{0.704}& \underline{0.727}&   \underline{0.695}&\underline{0.748}&\underline{0.728} & \textbf{1.889}&  \underline{2.025}& \underline{1.490}&\underline{1.575}\\
\bottomrule
\end{tabular}
\end{adjustbox}
\label{tab:dst_compare}
\end{table}

\begin{figure}[htbp]
	\centering
	{\includegraphics[width=\linewidth]{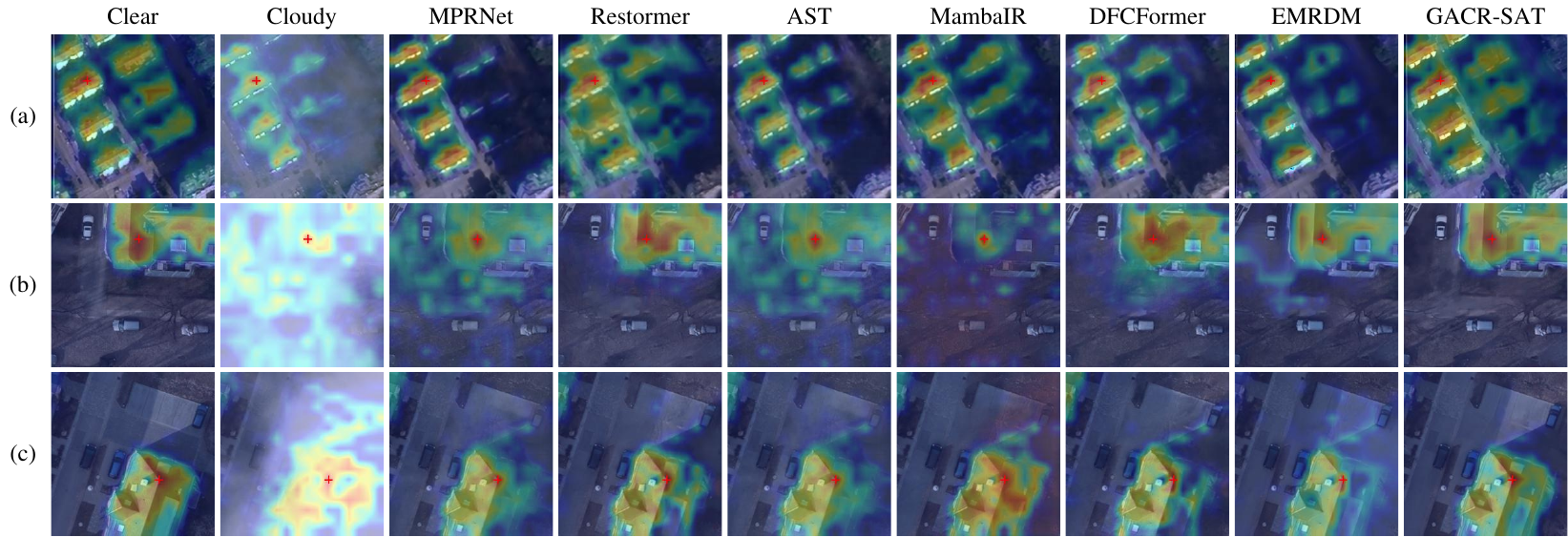}}
	\caption{Heatmaps of different CR obtained from the pretrained DINOv3 ViT-L/16-LVD-1689M. Regions with higher intensity indicate stronger similarity to the locations marked by red crosses.}
	\label{fig:cam}
\end{figure}

\subsection{Downstream Evaluation}
\label{subsec:dstev}

This section evaluates downstream performance to examine whether GACR preserves task-relevant semantic structures beyond pixel-level fidelity. The quantitative results in \cref{tab:dst_compare} are obtained using a DINOv3-based downstream encoder initialized with the ViT-L/16-LVD-1689M weights. In the table, the upper bound denotes testing directly on clear images, the lower bound represents performance without CR preprocessing, and end-to-end refers to training and testing downstream models directly on cloudy images. Results obtained using the ViT-L/16-SAT-300M weights are provided in Appendix~E.

Across all downstream tasks, GACR consistently achieves superior downstream performance compared with existing CR methods. In particular, GACR-SAT/2 attains the highest classification accuracies of 0.833 and 0.781 on CLS-1 and CLS-2, exceeding the strongest baseline (AST) by 2.0\% and 1.4\%, respectively. For height estimation, GACR-SAT/2 achieves the lowest RMSE values of 1.891 and 1.482 on HE-1 and HE-3, outperforming the second-best method by 0.1-0.2. These improvements indicate that GACR better preserves semantic consistency and spatial structure critical for downstream recognition.

For the CLS task, the performance gap between different CR methods is relatively small. We observe that the end-to-end approach yields results close to the upper bound and, in some cases, performs comparably to inputs preprocessed via CR. This phenomenon can be attributed to the global nature of classification, where coarse semantic representations are less sensitive to localized cloud occlusion. The t-SNE analysis in Appendix~E further supports this observation: cloudy inputs already form distinguishable clusters similar to cloud-free samples, implying that CR only provides limited additional separability.

\begin{figure}[htbp]
	\centering
	{\includegraphics[width=\linewidth]{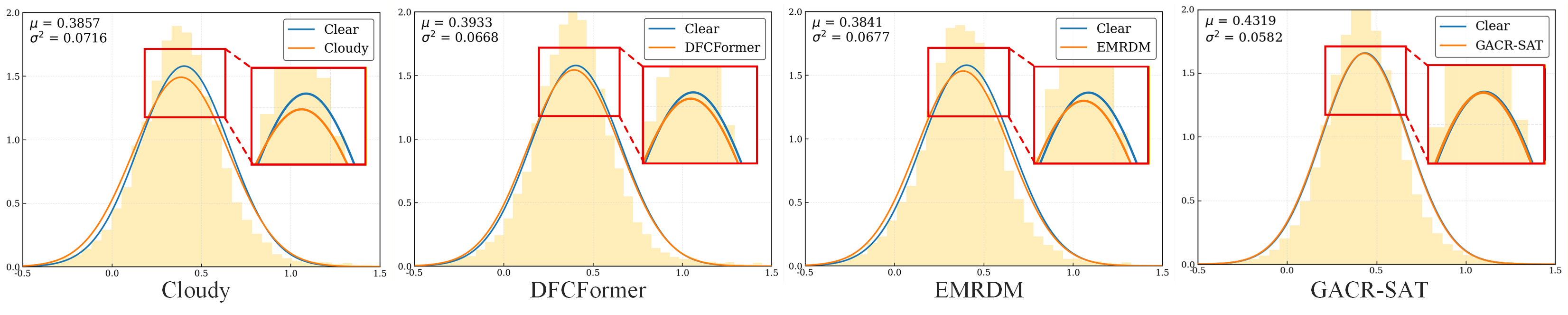}}
	\caption{Feature distance distribution comparison between CR result and corresponding cloud-free reference.}
	\label{fig:hist}
\end{figure}

In contrast, for dense prediction tasks such as segmentation and height estimation, CR leads to substantial downstream improvements, with GACR consistently delivering greater gains than competing methods. As shown in \cref{fig:viz_compare}, different CR approaches enhance land-cover discriminability to varying extents, yet GACR produces more distinct and semantically coherent object boundaries. To further analyze this behavior, we present feature activation maps in \cref{fig:cam}. For example, in row (a), GACR enables clearer separation of cloud-affected buildings on the right side of the image, indicating improved semantic localization. Moreover, \cref{fig:hist} visualizes feature distance distributions across methods. While cloudy inputs exhibit a noticeable distribution shift relative to clear references, GACR more effectively aligns the feature distribution with that of the cloud-free references. This observation is consistent with the geo-contextual alignment mechanism, which constrains reconstruction within a semantically coherent feature manifold.

\subsection{Ablation Study}
\label{subsec:abla}

\begin{wrapfigure}{r}{0.54\textwidth}
\vspace{-24pt}
\centering
\includegraphics[width=\linewidth]{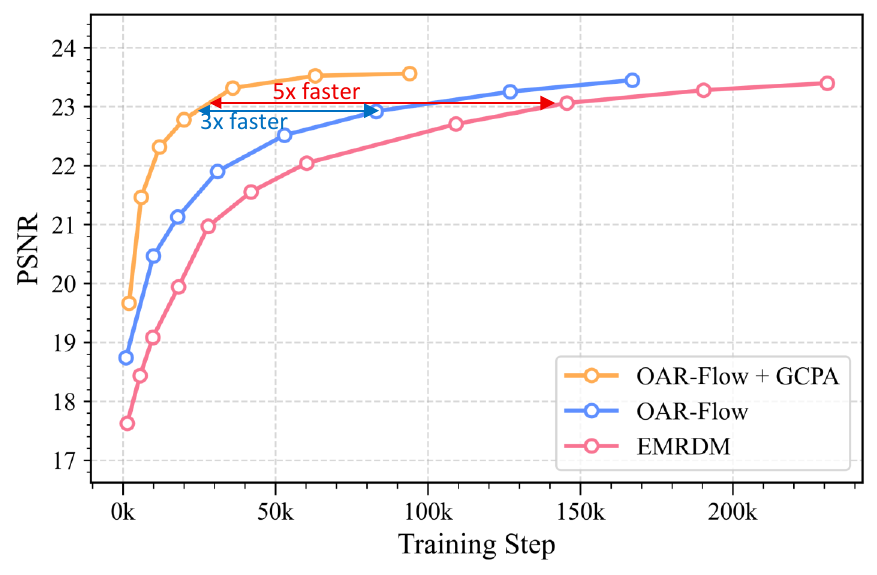}
\caption{The introduction of OAR-Flow and GCPA significantly accelerates the convergence of training.}
\label{fig:step}
\vspace{-24pt}
\end{wrapfigure}

\noindent\textbf{Convergence Speed.}
\cref{fig:step} presents the PSNR progression over training steps for EMRDM, OAR-Flow without GCPA, and the complete GACR model. Compared with EMRDM, OAR-Flow converges substantially faster, reaching the high-PSNR regime with significantly fewer iterations. Specifically, OAR-Flow 
achieves a comparable PSNR level using approximately one-third of the training steps required by EMRDM, corresponding to about a 3$\times$ acceleration in convergence. This improvement highlights the optimization efficiency introduced by the observation-anchored residual trajectory. Furthermore, incorporating GCPA further accelerates convergence. The complete GACR not only attains higher PSNR but also converges approximately 5$\times$ faster than EMRDM.

\begin{table}[htb]
\centering
\caption{Comparison of different model settings on CUHKCR-EXT-CS.}
\begin{tabular}{l|ccccc}
\toprule
Method & PSNR $\uparrow$& SSIM $\uparrow$& CLS $\uparrow$& BLD $\uparrow$ & GFLOPs\\
\midrule
DiT + MRDM& \underline{24.037}& 0.707 & 0.670 & 0.693 
 &697.20 
\\
HDiT + MRDM & 23.736 & \textbf{0.712} & \underline{0.726}& 0.696 
 &166.72 
\\
HDiT + OAR-Flow & 23.986 & 0.698 & 0.690 & \underline{0.700} &56.05 
\\
HDiT + OAR-Flow + GCPA & \textbf{24.230} & \underline{0.709}& \textbf{0.781} & \textbf{0.710} 
 &56.05 
\\
\bottomrule
\end{tabular}
\label{tab:model_comparison}
\end{table}

\noindent\textbf{Effectiveness of Model Choice.}
To investigate the effectiveness of each module, we perform an ablation study on the CUHKCR-EXT-CS dataset, as reported in \cref{tab:model_comparison}. Four configurations are compared to disentangle the effects of generative formulation and geo-contextual alignment. The diffusion-based baseline corresponding to EMRDM serves as a reference, while the complete GACR represents the final configuration. Although DiT-based modeling produces competitive visual results, it incurs substantially higher computational cost (GFLOPs = 697.20). Replacing diffusion dynamics with OAR-Flow consistently improves reconstruction metrics while maintaining efficiency. Moreover, incorporating GCPA further enhances both CR quality and downstream performance, indicating that geo-contextual alignment effectively mitigates semantic drift and strengthens interpretation reliability.

\begin{wraptable}{r}{0.58\textwidth}
\vspace{-34pt}
\centering
\caption{Hyperparameter analysis on Vaihingen-CR-Thick.}
\label{tab:patch_step}
\setlength{\tabcolsep}{2.2pt}
\renewcommand{\arraystretch}{0.92}
\small
\begin{tabular}{c|ccccc}
\toprule
  & PSNR $\uparrow$& SSIM $\uparrow$& SEG $\uparrow$& HE $\downarrow$ & GFLOPs\\
\midrule
   $\lambda$ & \multicolumn{5}{c}{$p=2$}\\
\midrule
  0.2 
&      \textbf{33.073}&       \textbf{0.965}&      0.728 &       1.578&-\\
                                            0.5
& 33.018 & 0.964 &      \textbf{0.737}&       \textbf{1.554}&-\\
                                            1.0 
&      32.838 &       0.964 &      0.730 &       1.563&-\\
                                            2.0 
&      32.720 &       0.963 &      0.729 &       1.576&-\\
                                            5.0 &      32.615 &       0.962 &      0.727 &       1.596&-\\
\midrule
   $p$ & \multicolumn{5}{c}{$\lambda=0.5$}\\
\midrule
  4 
& 30.303 & 0.949 &      0.693&       1.616&14.15 
\\
                                          2 
& 33.018 & 0.964 &      \textbf{0.737}&       1.554&56.05 
\\
                                          1 & \textbf{33.799} & \textbf{0.969} &      0.720 &       \textbf{1.548}&223.60 
\\
\bottomrule
\end{tabular}
\vspace{-10pt}
\end{wraptable}

\noindent\textbf{Effectiveness of $\mathbf{\lambda}$ and patch size $p$.}
We further conduct ablation studies on key hyperparameters in \cref{tab:patch_step}, including the balancing coefficient $\lambda$ for GCI and the patch size $p$. When $\lambda = 0.5$, the model achieves the best trade-off between reconstruction fidelity and downstream performance. This indicates that moderate regularization enhances semantic consistency without over-constraining pixel-level refinement, whereas overly small or large $\lambda$ values degrade either reconstruction quality or downstream metrics. Regarding patch size, smaller values lead to improved visual quality due to finer spatial modeling; however, setting $p = 1$ significantly increases computational cost in terms of FLOPs. Therefore, $p = 2$ is adopted as the default configuration in our experiments to balance reconstruction quality and computational efficiency while maintaining fair comparison with existing methods.

\section{Conclusion}

In this paper, we present GACR, an interpretation-oriented framework that jointly improves reconstruction fidelity and downstream reliability. With OAR-Flow, CR is reformulated as a physically grounded residual inversion process anchored to cloudy observations, reducing geographically implausible artifacts. Meanwhile, GCPA constrains reconstruction within a VFM-induced semantic manifold to preserve task-relevant structures and category-specific information. Extensive evaluations on six datasets and twelve downstream tasks show that GACR consistently enhances both reconstruction quality and task accuracy across diverse cloud conditions and remote sensing scenarios. These results suggest that coupling observation-anchored physical priors with semantic constraints enables CR to move beyond visual enhancement toward a reliable, semantics-preserving foundation for Earth observation.

\section*{Acknowledgements}
This work was supported by the Guangdong Science and Technology Department (Grant No. 2025A0505000062) and the Hong Kong Research Grants Council through the General Research Fund (Grant No. 17617024).

%
%
\bibliographystyle{splncs04}
\bibliography{main}

@ARTICLE{wang2025downstream,
  author={Wang, Ziyao and Ma, Xianping and Pun, Man-On},
  journal={IEEE Journal of Selected Topics in Applied Earth Observations and Remote Sensing}, 
  title={Downstream Task-Aware Cloud Removal for Very-High-Resolution Remote Sensing Images: An Information Loss Perspective}, 
  year={2025},
  volume={18},
  number={},
  pages={24531-24545},
}

@inproceedings{huh2024prh,
  title={The Platonic Representation Hypothesis},
  author={Huh, Minyoung and Cheung, Brian and Wang, Tongzhou and Isola, Phillip},
  booktitle={International Conference on Machine Learning},
  year={2024}
}

@inproceedings{crowson2024scalable,
  title={Scalable high-resolution pixel-space image synthesis with hourglass diffusion transformers},
  author={Crowson, Katherine and Baumann, Stefan Andreas and Birch, Alex and Abraham, Tanishq Mathew and Kaplan, Daniel Z and Shippole, Enrico},
  booktitle={International Conference on Machine Learning},
  year={2024}
}

@inproceedings{ma2024sit,
  title={{SiT}: Exploring flow and diffusion-based generative models with scalable interpolant transformers},
  author={Ma, Nanye and Goldstein, Mark and Albergo, Michael S and Boffi, Nicholas M and Vanden-Eijnden, Eric and Xie, Saining},
  booktitle={European Conference on Computer Vision},
  pages={23--40},
  year={2024},
  organization={Springer}
}

@inproceedings{liu2025effective,
  title={Effective cloud removal for remote sensing images by an improved mean-reverting denoising model with elucidated design space},
  author={Liu, Yi and Li, Wengen and Guan, Jihong and Zhou, Shuigeng and Zhang, Yichao},
  booktitle={Proceedings of the Computer Vision and Pattern Recognition Conference},
  pages={17851--17861},
  year={2025}
}

@article{luo2023image,
  title={Image Restoration with Mean-Reverting Stochastic Differential Equations},
  author={Luo, Ziwei and Gustafsson, Fredrik K and Zhao, Zheng and Sj{\"o}lund, Jens and Sch{\"o}n, Thomas B},
  journal={International Conference on Machine Learning},
  year={2023},
  organization={PMLR}
}

@misc{simeoni2025dinov3,
  title={{DINOv3}},
  author={Sim{\'e}oni, Oriane and Vo, Huy V and Seitzer, Maximilian and Baldassarre, Federico and Oquab, Maxime and Jose, Cijo and Khalidov, Vasil and Szafraniec, Marc and Yi, Seungeun and Ramamonjisoa, Micha{\"e}l and others},
  year={2025},
  eprint={2508.10104},
  archivePrefix={arXiv},
  primaryClass={cs.CV},
}

@article{zhu2017deep,
  title={Deep learning in remote sensing: A comprehensive review and list of resources},
  author={Zhu, Xiao Xiang and Tuia, Devis and Mou, Lichao and Xia, Gui-Song and Zhang, Liangpei and Xu, Feng and Fraundorfer, Friedrich},
  journal={IEEE geoscience and remote sensing magazine},
  volume={5},
  number={4},
  pages={8--36},
  year={2017},
  publisher={IEEE}
}

@article{li2022cloud,
  title={Cloud and cloud shadow detection for optical satellite imagery: Features, algorithms, validation, and prospects},
  author={Li, Zhiwei and Shen, Huanfeng and Weng, Qihao and Zhang, Yuzhuo and Dou, Peng and Zhang, Liangpei},
  journal={ISPRS Journal of Photogrammetry and Remote Sensing},
  volume={188},
  pages={89--108},
  year={2022},
  publisher={Elsevier}
}

@article{ebel2022sen12ms,
  title={{SEN12MS-CR-TS}: A remote-sensing data set for multimodal multitemporal cloud removal},
  author={Ebel, Patrick and Xu, Yajin and Schmitt, Michael and Zhu, Xiao Xiang},
  journal={IEEE Transactions on Geoscience and Remote Sensing},
  volume={60},
  pages={1--14},
  year={2022},
  publisher={IEEE}
}

@article{sui2024diffusion,
  author={Sui, Jialu and Ma, Yiyang and Yang, Wenhan and Zhang, Xiaokang and Pun, Man-On and Liu, Jiaying},
  journal={IEEE Transactions on Geoscience and Remote Sensing}, 
  title={Diffusion Enhancement for Cloud Removal in Ultra-Resolution Remote Sensing Imagery}, 
  year={2024},
  volume={62},
  pages={1-14},
}

@article{zou2024diffcr,
  title={{DiffCR}: A fast conditional diffusion framework for cloud removal from optical satellite images},
  author={Zou, Xuechao and Li, Kai and Xing, Junliang and Zhang, Yu and Wang, Shiying and Jin, Lei and Tao, Pin},
  journal={IEEE Transactions on Geoscience and Remote Sensing},
  volume={62},
  pages={1--14},
  year={2024},
  publisher={IEEE}
}

@article{bermudez2018sar,
  title={{SAR} to optical image synthesis for cloud removal with generative adversarial networks},
  author={Bermudez, Jose D and Happ, Patrick Nigri and Oliveira, Dario Augusto Borges and Feitosa, Raul Queiroz},
  journal={ISPRS Annals of the Photogrammetry, Remote Sensing and Spatial Information Sciences},
  volume={4},
  pages={5--11},
  year={2018},
  publisher={Copernicus GmbH}
}

@inproceedings{enomoto2017filmy,
  title={Filmy cloud removal on satellite imagery with multispectral conditional generative adversarial nets},
  author={Enomoto, Kenji and Sakurada, Ken and Wang, Weimin and Fukui, Hiroshi and Matsuoka, Masashi and Nakamura, Ryosuke and Kawaguchi, Nobuo},
  booktitle={Proceedings of the IEEE conference on computer vision and pattern recognition workshops},
  pages={48--56},
  year={2017}
}

@inproceedings{grohnfeldt2018conditional,
  title={A conditional generative adversarial network to fuse {SAR} and multispectral optical data for cloud removal from {Sentinel-2} images},
  author={Grohnfeldt, Claas and Schmitt, Michael and Zhu, Xiaoxiang},
  booktitle={IGARSS 2018-2018 IEEE International Geoscience and Remote Sensing Symposium},
  pages={1726--1729},
  year={2018},
  organization={IEEE}
}

@article{ma2023cloud,
  title={{Cloud-EGAN}: Rethinking cyclegan from a feature enhancement perspective for cloud removal by combining cnn and transformer},
  author={Ma, Xianping and Huang, Yiming and Zhang, Xiaokang and Pun, Man-On and Huang, Bo},
  journal={IEEE Journal of Selected Topics in Applied Earth Observations and Remote Sensing},
  volume={16},
  pages={4999--5012},
  year={2023},
  publisher={IEEE}
}

@article{li2019thick,
  title={Thick cloud removal with optical and {SAR} imagery via convolutional-mapping-deconvolutional network},
  author={Li, Wenbo and Li, Ying and Chan, Jonathan Cheung-Wai},
  journal={IEEE Transactions on Geoscience and Remote Sensing},
  volume={58},
  number={4},
  pages={2865--2879},
  year={2019},
  publisher={IEEE}
}

@article{zi2021thin,
  title={Thin cloud removal for multispectral remote sensing images using convolutional neural networks combined with an imaging model},
  author={Zi, Yue and Xie, Fengying and Zhang, Ning and Jiang, Zhiguo and Zhu, Wentao and Zhang, Haopeng},
  journal={IEEE Journal of Selected Topics in Applied Earth Observations and Remote Sensing},
  volume={14},
  pages={3811--3823},
  year={2021},
  publisher={IEEE}
}

@inproceedings{singh2018cloud,
  title={{Cloud-Gan}: Cloud Removal for Sentinel-2 Imagery Using a Cyclic Consistent Generative Adversarial Networks},
  author={Singh, Praveer and Komodakis, Nikos},
  booktitle={IGARSS 2018-2018 IEEE International Geoscience and Remote Sensing Symposium},
  pages={1772--1775},
  year={2018},
  organization={IEEE}
}

@article{jin2024rfe,
  title={{RFE-VCR}: Reference-enhanced transformer for remote sensing video cloud removal},
  author={Jin, Xianyu and He, Jiang and Xiao, Yi and Lihe, Ziyang and Liao, Xusi and Li, Jie and Yuan, Qiangqiang},
  journal={ISPRS Journal of Photogrammetry and Remote Sensing},
  volume={214},
  pages={179--192},
  year={2024},
  publisher={Elsevier}
}

@article{liu2025cr,
  title={{CR-Famba}: A frequency-domain assisted mamba for thin cloud removal in optical remote sensing imagery},
  author={Liu, Jiao and Pan, Bin and Shi, Zhenwei},
  journal={IEEE Transactions on Multimedia},
  year={2025},
  publisher={IEEE}
}

@article{pan2025m,
  title={{M3-CR}: Multi-scale Multi-branch {Mamba} for {SAR}-assisted Optical Image Thick Cloud Removal},
  author={Pan, Linpeng and Song, Xuedong and Xie, Fengying and Zhang, Xiaozhe and Ji, Haolin and Shi, Zhenwei},
  journal={IEEE Transactions on Geoscience and Remote Sensing},
  year={2025},
  publisher={IEEE}
}

@article{silva2025cloud,
  title={Cloud removal with compact diffusion models: A residual block-based approach},
  author={Silva, Leandro Henrique Furtado Pinto and Mari, Jo{\~a}o Fernando and Escarpinati, Mauricio Cunha and Backes, Andr{\'e} Ricardo},
  journal={Remote Sensing Applications: Society and Environment},
  pages={101680},
  year={2025},
  publisher={Elsevier}
}

@inproceedings{chen2025unirestore,
  title={Unirestore: Unified perceptual and task-oriented image restoration model using diffusion prior},
  author={Chen, I and Chen, Wei-Ting and Liu, Yu-Wei and Chiang, Yuan-Chun and Kuo, Sy-Yen and Yang, Ming-Hsuan and others},
  booktitle={Proceedings of the Computer Vision and Pattern Recognition Conference},
  pages={17969--17979},
  year={2025}
}

@inproceedings{peebles2023scalable,
  title={Scalable diffusion models with transformers},
  author={Peebles, William and Xie, Saining},
  booktitle={Proceedings of the IEEE/CVF international conference on computer vision},
  pages={4195--4205},
  year={2023}
}

@inproceedings{rombach2022high,
  title={High-resolution image synthesis with latent diffusion models},
  author={Rombach, Robin and Blattmann, Andreas and Lorenz, Dominik and Esser, Patrick and Ommer, Bj{\"o}rn},
  booktitle={Proceedings of the IEEE/CVF conference on computer vision and pattern recognition},
  pages={10684--10695},
  year={2022}
}

@article{ho2020denoising,
  title={Denoising diffusion probabilistic models},
  author={Ho, Jonathan and Jain, Ajay and Abbeel, Pieter},
  journal={Advances in neural information processing systems},
  volume={33},
  pages={6840--6851},
  year={2020}
}

@inproceedings{nichol2021improved,
  title={Improved denoising diffusion probabilistic models},
  author={Nichol, Alexander Quinn and Dhariwal, Prafulla},
  booktitle={International conference on machine learning},
  pages={8162--8171},
  year={2021},
  organization={PMLR}
}

@article{song2020denoising,
  title={Denoising Diffusion Implicit Models},
  author={Song, Jiaming and Meng, Chenlin and Ermon, Stefano},
  journal={arXiv:2010.02502},
  year={2020},
  month={October},
  abbr={Preprint},
}

@inproceedings{song2021scorebased,
  title={Score-Based Generative Modeling through Stochastic Differential Equations},
  author={Yang Song and Jascha Sohl-Dickstein and Diederik P Kingma and Abhishek Kumar and Stefano Ermon and Ben Poole},
  booktitle={International Conference on Learning Representations},
  year={2021},
}

@article{goodfellow2020generative,
  title={Generative adversarial networks},
  author={Goodfellow, Ian and Pouget-Abadie, Jean and Mirza, Mehdi and Xu, Bing and Warde-Farley, David and Ozair, Sherjil and Courville, Aaron and Bengio, Yoshua},
  journal={Communications of the ACM},
  volume={63},
  number={11},
  pages={139--144},
  year={2020},
  publisher={ACM New York, NY, USA}
}

@article{kingma2013auto,
  title={Auto-encoding variational bayes},
  author={Kingma, Diederik P and Welling, Max},
  journal={arXiv preprint arXiv:1312.6114},
  year={2013}
}

@article{karras2022elucidating,
  title={Elucidating the design space of diffusion-based generative models},
  author={Karras, Tero and Aittala, Miika and Aila, Timo and Laine, Samuli},
  journal={Advances in neural information processing systems},
  volume={35},
  pages={26565--26577},
  year={2022}
}

@article{albergo2023stochastic,
  title={Stochastic interpolants: A unifying framework for flows and diffusions},
  author={Albergo, Michael S and Boffi, Nicholas M and Vanden-Eijnden, Eric},
  journal={arXiv preprint arXiv:2303.08797},
  year={2023}
}

@inproceedings{albergo2023building,
    title={Building Normalizing Flows with Stochastic Interpolants},
    author={Michael Samuel Albergo and Eric Vanden-Eijnden},
    booktitle={The Eleventh International Conference on Learning Representations },
    year={2023},
}

@inproceedings{liu2022,
    title={Flow Straight and Fast: Learning to Generate and Transfer Data with Rectified Flow},
    author={Xingchao Liu and Chengyue Gong and Qiang Liu},
    booktitle={The Eleventh International Conference on Learning Representations },
    year={2023},
}

@inproceedings{lipman2022,
    title={Flow Matching for Generative Modeling},
    author={Yaron Lipman and Ricky T. Q. Chen and Heli Ben-Hamu and Maximilian Nickel and Matthew Le},
    booktitle={The Eleventh International Conference on Learning Representations },
    year={2023},
}

@article{lu2022dpm,
  title={{DPM}-solver: A fast {ODE} solver for diffusion probabilistic model sampling in around 10 steps},
  author={Lu, Cheng and Zhou, Yuhao and Bao, Fan and Chen, Jianfei and Li, Chongxuan and Zhu, Jun},
  journal={Advances in neural information processing systems},
  volume={35},
  pages={5775--5787},
  year={2022}
}

@article{de2022riemannian,
  title={Riemannian score-based generative modelling},
  author={De Bortoli, Valentin and Mathieu, Emile and Hutchinson, Michael and Thornton, James and Teh, Yee Whye and Doucet, Arnaud},
  journal={Advances in neural information processing systems},
  volume={35},
  pages={2406--2422},
  year={2022}
}

@article{bechtel2020weighted,
  title={A Weighted Accuracy Measure for Land Cover Mapping: Comment on Johnson et al. Local Climate Zone ({LCZ}) Map Accuracy Assessments Should Account for Land Cover Physical Characteristics that Affect the Local Thermal Environment. Remote Sens. 2019, 11, 2420},
  author={Bechtel, Benjamin and Demuzere, Matthias and Stewart, Iain D},
  journal={Remote Sensing},
  volume={12},
  number={11},
  pages={1769},
  year={2020},
  publisher={MDPI}
}

@article{czerkawski2023satellitecloudgenerator,
  title={Satellitecloudgenerator: controllable cloud and shadow synthesis for multi-spectral optical satellite images},
  author={Czerkawski, Mikolaj and Atkinson, Robert and Michie, Craig and Tachtatzis, Christos},
  journal={Remote Sensing},
  volume={15},
  number={17},
  pages={4138},
  year={2023},
  publisher={MDPI}
}

@article{he2010single,
  title={Single image haze removal using dark channel prior},
  author={He, Kaiming and Sun, Jian and Tang, Xiaoou},
  journal={IEEE transactions on pattern analysis and machine intelligence},
  volume={33},
  number={12},
  pages={2341--2353},
  year={2010},
  publisher={IEEE}
}

@inproceedings{mehri2021mprnet,
  title={{MPRNet}: Multi-Path Residual Network for Lightweight Image Super Resolution},
  author={Mehri, Armin and Ardakani, Parichehr B and Sappa, Angel D},
  booktitle={Proceedings of the IEEE/CVF Winter Conference on Applications of Computer Vision},
  pages={2704--2713},
  year={2021}
}

@inproceedings{zamir2022restormer,
  title={Restormer: Efficient Transformer for High-Resolution Image Restoration},
  author={Zamir, Syed Waqas and Arora, Aditya and Khan, Salman and Hayat, Munawar and Khan, Fahad Shahbaz and Yang, Ming-Hsuan},
  booktitle={Proceedings of the IEEE/CVF conference on computer vision and pattern recognition},
  pages={5728--5739},
  year={2022}
}

@inproceedings{zhou2024adapt,
  title={Adapt or perish: Adaptive sparse transformer with attentive feature refinement for image restoration},
  author={Zhou, Shihao and Chen, Duosheng and Pan, Jinshan and Shi, Jinglei and Yang, Jufeng},
  booktitle={Proceedings of the IEEE/CVF Conference on Computer Vision and Pattern Recognition},
  pages={2952--2963},
  year={2024}
}

@inproceedings{guo2025mambair,
  title={{MambaIR}: A simple baseline for image restoration with state-space model},
  author={Guo, Hang and Li, Jinmin and Dai, Tao and Ouyang, Zhihao and Ren, Xudong and Xia, Shu-Tao},
  booktitle={European Conference on Computer Vision},
  pages={222--241},
  year={2024},
  organization={Springer}
}

@inproceedings{gu2025acl,
  title={{ACL}: Activating Capability of Linear Attention for Image Restoration},
  author={Gu, Yubin and Meng, Yuan and Ji, Jiayi and Sun, Xiaoshuai},
  booktitle={Proceedings of the Computer Vision and Pattern Recognition Conference},
  pages={17913--17923},
  year={2025}
}

@inproceedings{wang2025reconciling,
  title={Reconciling Stochastic and Deterministic Strategies for Zero-shot Image Restoration using Diffusion Model in Dual},
  author={Wang, Chong and Guo, Lanqing and Fu, Zixuan and Yang, Siyuan and Cheng, Hao and Kot, Alex C and Wen, Bihan},
  booktitle={Proceedings of the Computer Vision and Pattern Recognition Conference},
  pages={23207--23216},
  year={2025}
}

@inproceedings{jeong2025latent,
  title={Latent space super-resolution for higher-resolution image generation with diffusion models},
  author={Jeong, Jinho and Han, Sangmin and Kim, Jinwoo and Kim, Seon Joo},
  booktitle={Proceedings of the Computer Vision and Pattern Recognition Conference},
  pages={2355--2365},
  year={2025}
}

@inproceedings{yang2025fam,
  title={{FAM} diffusion: Frequency and attention modulation for high-resolution image generation with stable diffusion},
  author={Yang, Haosen and Bulat, Adrian and Hadji, Isma and Pham, Hai X and Zhu, Xiatian and Tzimiropoulos, Georgios and Martinez, Brais},
  booktitle={Proceedings of the Computer Vision and Pattern Recognition Conference},
  pages={2459--2468},
  year={2025}
}

@inproceedings{feng2025gps,
  title={{GPS} as a Control Signal for Image Generation},
  author={Feng, Chao and Chen, Ziyang and Holynski, Aleksander and Efros, Alexei A and Owens, Andrew},
  booktitle={Proceedings of the Computer Vision and Pattern Recognition Conference},
  pages={2766--2778},
  year={2025}
}

@inproceedings{astruc2024omnisat,
  title={Omnisat: Self-supervised modality fusion for earth observation},
  author={Astruc, Guillaume and Gonthier, Nicolas and Mallet, Clement and Landrieu, Loic},
  booktitle={European Conference on Computer Vision},
  pages={409--427},
  year={2024},
  organization={Springer}
}

@inproceedings{guo2024skysense,
  title={Skysense: A multi-modal remote sensing foundation model towards universal interpretation for earth observation imagery},
  author={Guo, Xin and Lao, Jiangwei and Dang, Bo and Zhang, Yingying and Yu, Lei and Ru, Lixiang and Zhong, Liheng and Huang, Ziyuan and Wu, Kang and Hu, Dingxiang and others},
  booktitle={Proceedings of the IEEE/CVF Conference on Computer Vision and Pattern Recognition},
  pages={27672--27683},
  year={2024}
}

@inproceedings{zhu2025skysense,
  title={{Skysense-O}: Towards open-world remote sensing interpretation with vision-centric visual-language modeling},
  author={Zhu, Qi and Lao, Jiangwei and Ji, Deyi and Luo, Junwei and Wu, Kang and Zhang, Yingying and Ru, Lixiang and Wang, Jian and Chen, Jingdong and Yang, Ming and others},
  booktitle={Proceedings of the Computer Vision and Pattern Recognition Conference},
  pages={14733--14744},
  year={2025}
}

@article{king2013spatial,
  title={Spatial and temporal distribution of clouds observed by MODIS onboard the Terra and Aqua satellites},
  author={King, Michael D and Platnick, Steven and Menzel, W Paul and Ackerman, Steven A and Hubanks, Paul A},
  journal={IEEE transactions on geoscience and remote sensing},
  volume={51},
  number={7},
  pages={3826--3852},
  year={2013},
  publisher={IEEE}
}

\clearpage
\onecolumn
\setcounter{page}{1}
{
    \newpage
    \centering
    \Large
    Supplementary Material \\
}
\appendix
\WarningFilter{latex}{Text page}

\section{Proof of the probability flow ODE with the velocity.}
\label{app:der}
In this part, we provide a detailed derivation of the backward process formulation introduced in the main paper. Specifically, we show how the marginal distribution $p_t(x)$ of variable $x_t$ in \cref{eq:oar_transport} satisfies the transport equation and how this leads to the expression of the velocity field $\mathbf{v}_t(x)$ in \cref{eq:oar_velocity}. Part of proofs are derived from \cite{albergo2023stochastic}.

Consider the time-dependent probability density function $p_t(x)$ of $x_t$ defined in \cref{eq:oar_forward}. By definition, its characteristic function $\hat{p}_t(\mathbf{k}) = \int_{\mathbb{R}^{d}} e^{i \mathbf{k} \cdot x} p_{t}(x) \mathrm{d}x$ is given by:
\begin{equation}
    \hat{p}_t(\mathbf{k}) = \mathbb{E}\!\left[e^{i\mathbf{k} \cdot x_t}\right], 
    \label{eq:char_func}
\end{equation}
where $\mathbb{E}$ denotes expectation over $x_*$, $x_c$ and $\epsilon$. Taking time derivation on both sides, and using the tower property of conditional expectation, we have:
\begin{align}
    \partial_t \hat{p}_t(\mathbf{k})
    &= i\,\mathbf{k}\cdot \mathbb{E}\!\left[\dot{x}_t \, e^{i\mathbf{k}\cdot x_t}\right]
    \label{eq:time_derivative1}\\
    &= i\,\mathbf{k}\cdot \mathbb{E}_{x\sim p_t}\!\left[
          \mathbb{E}\!\left[\dot{x}_t \, e^{i\mathbf{k}\cdot x_t}\,\middle|\, x_t=x\right]\right]
    \label{eq:time_derivative2}\\
    &= i\,\mathbf{k}\cdot \mathbb{E}_{x\sim p_t}\!\left[
          \mathbb{E}\!\left[(\dot{\alpha}_t x_*+\dot{\beta}_t x_c+\dot{\sigma}_t \epsilon)\,
          e^{i\mathbf{k}\cdot x_t}\,\middle|\, x_t=x\right]\right]
    \label{eq:time_derivative3}\\
    &= i\,\mathbf{k}\cdot \mathbb{E}_{x\sim p_t}\!\left[
          \mathbb{E}\!\left[(\dot{\alpha}_t x_*+\dot{\beta}_t x_c+\dot{\sigma}_t \epsilon)\,
          \middle|\, x_t=x\right] e^{i\mathbf{k}\cdot x}\right]
    \label{eq:time_derivative4}\\
    &= i\,\mathbf{k}\cdot \mathbb{E}_{x\sim p_t}\!\left[
          \mathbf{v}_t(x)\, e^{i\mathbf{k}\cdot x}\right]
    \label{eq:time_derivative5}
\end{align}
where $\mathbf{v}_t(x) = \mathbb{E}\!\left[(\dot{\alpha}_t x_* + \dot{\beta}_t x_c + \dot{\sigma}_t \epsilon) \mid x_t = x\right] = \dot{\alpha}_t \mathbb{E}[x_* \mid x_t = x] + \dot{\beta}_t \mathbb{E}[x_c \mid x_t = x] + \dot{\sigma}_t \mathbb{E}[\epsilon \mid x_t = x]$ is the velocity defined in \cref{eq:oar_velocity}. Furthermore, following \cref{eq:time_derivative5}, we have:
\begin{equation}
    \partial_t \int_{\mathbb{R}^d} e^{i\mathbf{k} \cdot x} p_t(x) \mathrm{d}x
    = i\mathbf{k} \cdot \int_{\mathbb{R}^d} \mathbf{v}_t(x) e^{i\mathbf{k} \cdot x} p_t(x) \mathrm{d}x, 
    \label{eq:fourier_transform}
\end{equation}
from which we deduce:
\begin{align}
    \int_{\mathbb{R}^d} e^{i\mathbf{k} \cdot x} \partial_t p_t(x) dx
    &= \int_{\mathbb{R}^d} \mathbf{v}_t(x) \cdot \nabla_{x}[e^{i\mathbf{k} \cdot x}] p_t(x) \mathrm{d}x
    \label{eq:integration_by_parts_1}\\[4pt]
    &= - \int_{\mathbb{R}^d} \nabla_{x} \cdot [\mathbf{v}_t(x) p_t(x)] e^{i\mathbf{k} \cdot x} \mathrm{d}x,
    \label{eq:integration_by_parts_2}
\end{align}
where $\nabla_{x} \cdot [\mathbf{v}p_t] = \sum_{i=1}^d \frac{\partial}{\partial x_i} [v_i p_t]$ is the divergence operator.  
By properties of the Fourier transform, this implies that $p_t(\mathbf{x})$ satisfies the transport equation in \cref{eq:oar_transport}.

\section{Downstream Networks}
\label{app:dst}

\begin{figure*}[htbp]
	\centering
	{\includegraphics[width=0.7\linewidth]{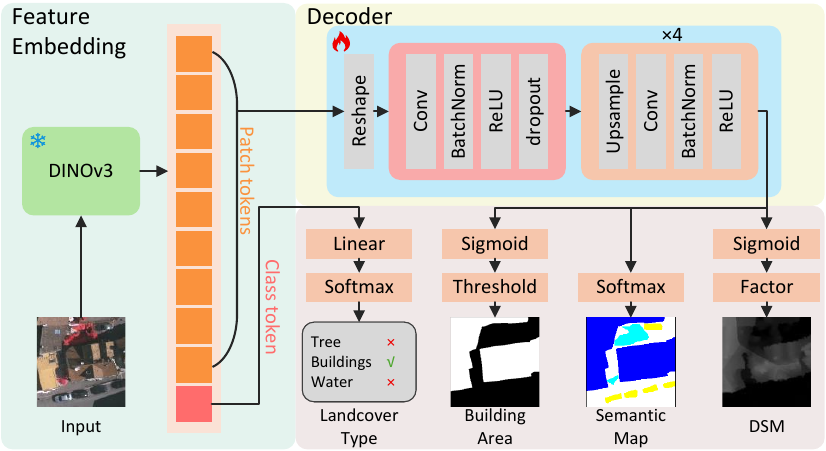}}
	\caption{Downstream Networks.}
	\label{fig:decoder}
\end{figure*}

Owing to the strong transferability and generalization capability of DINOv3 across various visual tasks, we adopt the frozen DINOv3 weights as the backbone of downstream networks. Only lightweight task-specific decoders are trained, as illustrated in \cref{fig:decoder}. In our implementation, DINOv3 produces both a class token and a set of patch tokens. The class token is used for the CLS task, while the reshaped patch tokens are employed for the three dense prediction tasks, including BLD, SEG, and HE, whose specific designs are described as follows.

For the CLS task, we adopt a simple yet effective strategy by directly applying a linear classifier to the class token. The classifier transforms the global representation into category probabilities, which are optimized using the standard cross-entropy loss.

For the three dense prediction tasks, we employ a unified lightweight cascaded decoder to progressively reconstruct spatial resolution from the patch features. Each stage consists of interpolation-based upsampling followed by convolution, activation, and normalization operations, enabling hierarchical refinement of spatial features until the original image scale is restored. A task-specific prediction head is then applied to generate the final output map.

For the BLD task, the prediction head outputs a single-channel probability map, activated by a sigmoid function and optimized using the binary cross-entropy loss. During inference, a fixed threshold of 0.5 is applied to binarize the output into building and non-building regions.
For the SEG task, a softmax activation is applied to produce multi-class probability maps, and the network is trained using a pixel-wise cross-entropy loss to ensure accurate semantic labeling.
For the HE task, the decoder outputs a normalized height map activated by a sigmoid function, which is optimized using an L1 loss. During evaluation, the predicted normalized heights are rescaled by a constant factor (25.5 in practice) to recover metric height values for quantitative assessment.

\section{Datasets Details}
\label{app:dataset}
\subsection{CUHKCR-EXT Dataset}

\begin{figure*}[ht]
	\centering
	{\includegraphics[width=0.60\linewidth]{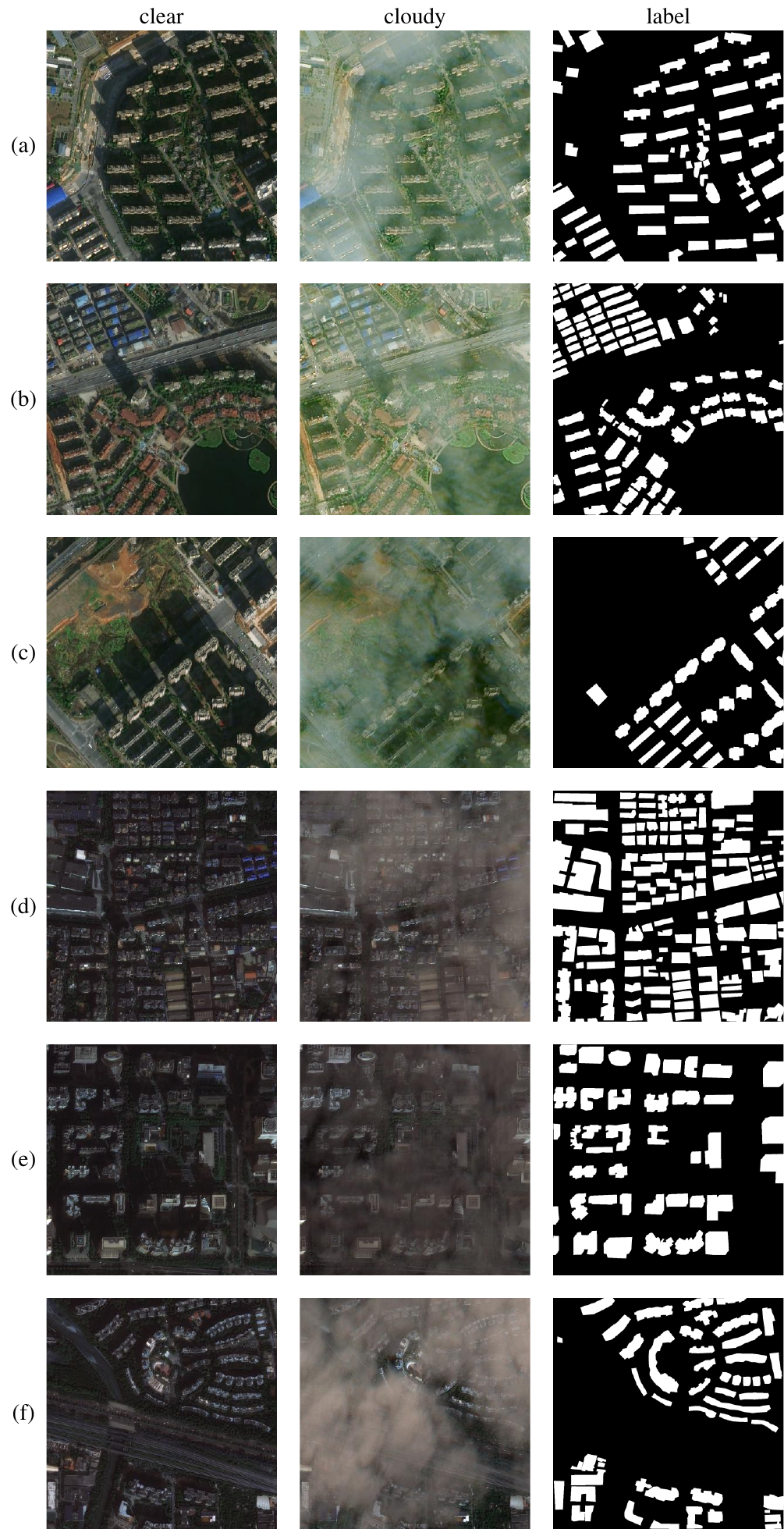}}
	\caption{Sample visualization for the BLD task. From left to right are the clear image, cloudy image, and building area label. Panels (a-c) are selected from the CUHKCR-EXT-CS dataset, while panels (d-f) are taken from the CUHKCR-EXT-GZ dataset.}
	\label{fig:viz_cuhk}
\end{figure*}

The CUHKCR-EXT dataset, introduced by \cite{wang2025downstream}, contains 0.5 m ultra-high-resolution images captured by the Jilin-1 satellite. It includes paired cloudy and cloud-free images from two regions in Guangzhou (GZ) and Changsha (CS), China. The CUHKCR-EXT-GZ subset is primarily covered by thin clouds, whereas the CUHKCR-EXT-CS subset exhibits a higher proportion of thick clouds. CUHKCR-EXT consists of two parts: the CR part and downstream tasks part. For the CR subset, CUHKCR-EXT-CS contains 34 large images of size 3000 $\times$ 3000, and CUHKCR-EXT-GZ includes 31 images of size 3600 $\times$ 3600. The downstream part provides annotations for six land-cover types based on the LCZ standard \cite{bechtel2020weighted}. Since the dataset provides unsliced, large-scale images, it enables flexible and customizable alignment for experimental comparison. To support our building extraction task, we manually annotated building regions using the CVAT tool.
The annotated regions used for building detection were strictly separated from those used for cloud removal to avoid overlap and ensure independent evaluation. Some samples used for BLD tasks are shown in \cref{fig:viz_cuhk}.

\subsection{Synthetic Cloud Dataset}

We trained and evaluated our downstream performance based on the ISPRS Vaihingen
and ISPRS Potsdam.
The Vaihingen and Potsdam datasets are two high-resolution benchmarks widely used for urban scene understanding. The Vaihingen dataset consists of 16 orthophotos (approximately 2500 $\times$ 2000 pixels each) with three spectral bands (near-infrared, red, and green) and a normalized DSM at a ground sampling distance of 9 cm. The Potsdam dataset contains 24 orthophotos of 6000 $\times$ 6000 pixels with four spectral bands (infrared, red, green, and blue) and a DSM at 5 cm resolution; the RGB channels are adopted in our experiments. They share six semantic categories, including building, tree, low vegetation, car, impervious surface, and a background class. 

\begin{figure*}[ht]
	\centering
	{\includegraphics[width=0.85\linewidth]{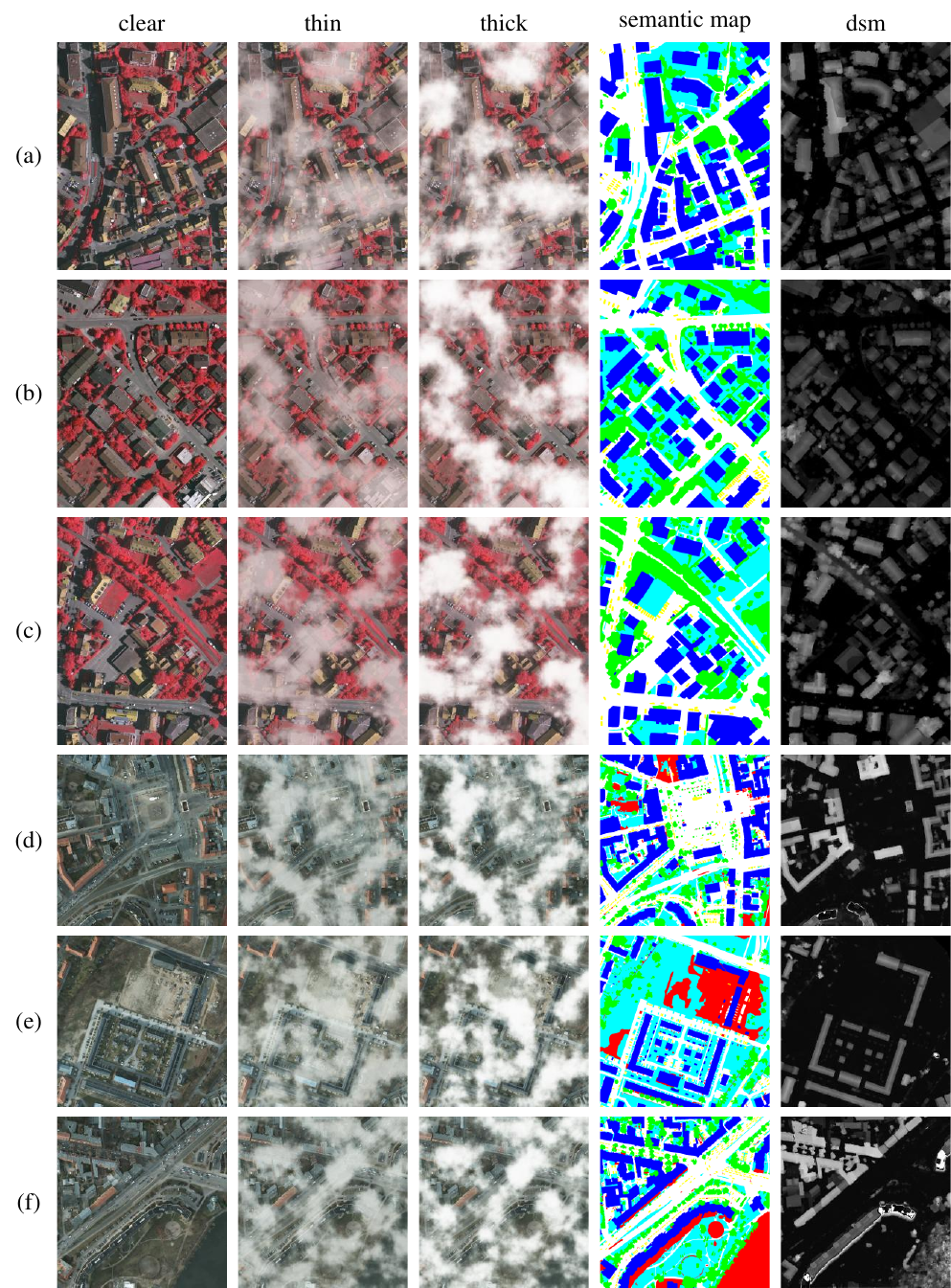}}
	\caption{Sample visualization for the SEG and HE tasks. From left to right are the clear image, thin-cloud image, thick-cloud image, semantic map, and DSM. Panels (a-c) are selected from the Vaihingen-CR-thin and Vaihingen-CR-thick datasets, while panels (d-f) are taken from the Vaihingen-CR-thick dataset.}
	\label{fig:viz_isprs}
\end{figure*}

To obtain high-resolution cloudy datasets with SEG and HE annotations, we constructed four cloud removal datasets containing both thin- and thick-cloud types based on the atmospheric scattering model \cite{he2010single} and the cloud generation approach proposed by Czerkawski et al. \cite{czerkawski2023satellitecloudgenerator}. To systematically control cloud opacity, we extended the cloud generator with a scalar thickness parameter that jointly scales the mask intensity. To ensure controlled variability, two constraints are imposed: (1) for a fixed thickness, different images exhibit distinct cloud shapes; and (2) within the same image, varying the thickness preserves the overall cloud structure. Sample visualizations are shown in \cref{fig:viz_isprs}.

\section{Detailed Experiment Implements}
\label{app:imple}

\subsection{Training Configurations}

To conduct our experiments, we trained GACR for 200k steps with a batch size of 4 and performed validation every 10k steps to select the model with the best performance. We adopt the AdamW optimizer with $\beta_1 = 0.9$, $\beta_2 = 0.999$, and $\epsilon = 1\times10^{-8}$, while the learning rate is set to $1\times10^{-4}$. Experiments using the DINOv3 ViT-L/16-SAT-300M weights with $p=1$ are conducted on a single A100 GPU, whereas all other reported results are obtained using a single A800 GPU. All images are resized to $256 \times 256$ for both training and inference. Additional training details are summarized in Table~\ref{tab:training_settings}.

\begin{table}[t]
\centering
\caption{Training configurations across datasets.}
\label{tab:training_settings}
\small
\renewcommand{\arraystretch}{1.1}
\setlength{\tabcolsep}{4pt}
\resizebox{\columnwidth}{!}{
\begin{tabular}{l|cccccc}
\toprule
Dataset & CUHKCR-EXT-GZ & CUHKCR-EXT-CS & Potsdam-CR-thin & Potsdam-CR-thick & Vaihingen-CR-thin & Vaihingen-CR-thick \\
\midrule
Training number& 3686& 2800& 3678& 3678& 1803&1803\\
Test number& 922& 700& 920& 920& 451&451\\
Image size & \multicolumn{6}{c}{256 $\times$ 256} \\
Training steps & \multicolumn{6}{c}{200k} \\
Batch size & \multicolumn{6}{c}{4} \\
Precision & \multicolumn{6}{c}{tf32} \\
PERA dim & \multicolumn{6}{c}{1024} \\
Training device & 1 $\times$ A800& 1 $\times$  A800& 1 $\times$  A800& 1 $\times$  A800& 1 $\times$  A800& 1 $\times$  A800\\
\midrule
Depth & \multicolumn{6}{c}{[2, 2, 2, 2]} \\
Widths & \multicolumn{6}{c}{[128, 256, 512, 1024]} \\
FFN intermediate widths & \multicolumn{6}{c}{[256, 512, 1024, 2048]} \\
Attention heads & \multicolumn{6}{c}{64} \\
Attention dim & \multicolumn{6}{c}{[2, 4, 6, 12]} \\
Neighborhood kernel size & \multicolumn{6}{c}{7} \\
Dropout rate & \multicolumn{6}{c}{[0.0, 0.0, 0.0, 0.1]} \\
\midrule
Optimizer & \multicolumn{6}{c}{AdamW} \\
Learning rate & \multicolumn{6}{c}{$1.0\times10^{-4}$} \\
$\mathrm{beta_1}$& \multicolumn{6}{c}{0.9} \\
$\mathrm{beta_2}$& \multicolumn{6}{c}{0.999} \\
$\mathrm{epsilon}$& \multicolumn{6}{c}{$1.0\times10^{-8}$} \\
Weight decay & \multicolumn{6}{c}{0} \\
$\lambda$ & \multicolumn{6}{c}{0.5} \\
Sampling steps & \multicolumn{6}{c}{4} \\
\bottomrule
\end{tabular}}
\end{table}

\subsection{Compared Methods}
We select six representative methods as baselines for comparison. Specifically, MPRNet \cite{mehri2021mprnet} is a CNN-based method; Restormer \cite{zamir2022restormer} and AST \cite{zhou2024adapt} are Transformer-based methods; MambaIR \cite{guo2025mambair} is Mamba-based; DFCFormer \cite{wang2025downstream} is dynamic-filter-based; and EMRDM \cite{liu2025effective} is diffusion-based. Although the PSNR and SSIM results of MPRNet, Restormer, AST, MambaIR, and DFCFormer were reported in \cite{wang2025downstream}, we retrain all these models to ensure fair comparison, since our experiments re-slice the original images to construct training sets that do not overlap with the BLD task. EMRDM is trained for 400k steps following its original configuration, while the other methods are trained for 100 epochs using their default settings. All reported results correspond to the checkpoints achieving the best validation performance.

\subsection{Evaluation Metrics}
\paragraph{Cloud Removal (CR).}
For the CR task, we evaluate the image restoration quality using four widely used metrics: Peak Signal-to-Noise Ratio (PSNR), Structural Similarity Index (SSIM), Learned Perceptual Image Patch Similarity (LPIPS), and Root Mean Squared Error (RMSE).  
The PSNR measures the reconstruction fidelity between the restored image $\hat{I}$ and the ground-truth clear image $I$, defined as:
\begin{equation}
\mathrm{PSNR} = 10 \log_{10} \left( \frac{L^2}{\frac{1}{N}\sum_{i=1}^{N}(I_i - \hat{I}_i)^2} \right),
\end{equation}
where $L$ is the maximum possible pixel value (e.g., 255 for 8-bit images), and $N$ denotes the total number of pixels.  
The SSIM measures the structural similarity between two images, formulated as:
\begin{equation}
\mathrm{SSIM}(I, \hat{I}) = 
\frac{(2\mu_I \mu_{\hat{I}} + C_1)(2\sigma_{I\hat{I}} + C_2)}
{(\mu_I^2 + \mu_{\hat{I}}^2 + C_1)(\sigma_I^2 + \sigma_{\hat{I}}^2 + C_2)},
\end{equation}
where $\mu$ and $\sigma$ denote mean and standard deviation, $\sigma_{I\hat{I}}$ represents cross-covariance, and $C_1$, $C_2$ are small constants to stabilize the division.  

LPIPS measures the perceptual distance between two images based on deep feature representations extracted by pretrained networks, and smaller LPIPS values indicate better perceptual quality, which is defined as:
\begin{equation}
\mathrm{LPIPS}(I, \hat{I}) = \sum_{l} \frac{1}{H_l W_l} 
\sum_{h=1}^{H_l} \sum_{w=1}^{W_l} 
\left\| w_l \odot \left( \phi_l(I)_{hw} - \phi_l(\hat{I})_{hw} \right) \right\|_2^2,
\end{equation}
where $\phi_l(\cdot)$ denotes the feature map of layer $l$ from the pretrained network, 
$H_l$ and $W_l$ are its spatial dimensions, 
$w_l$ is a learned weight vector that calibrates channel-wise importance, 
and $\odot$ represents element-wise multiplication.  

The RMSE quantifies pixel-wise error:
\begin{equation}
\mathrm{RMSE} = \sqrt{\frac{1}{N}\sum_{i=1}^{N}(I_i - \hat{I}_i)^2}.
\end{equation}

\paragraph{Classification (CLS).}
For the classification task, the overall accuracy (Acc) is used to evaluate model performance:
\begin{equation}
\mathrm{Acc} = \frac{N_{\text{correct}}}{N_{\text{total}}},
\end{equation}
where $N_{\text{correct}}$ is the number of correctly classified samples and $N_{\text{total}}$ is the total number of samples.

\paragraph{Building Detection (BLD).}
For the building detection task, the Intersection over Union (IoU) is adopted to measure the overlap between predicted and ground-truth building masks:
\begin{equation}
\mathrm{IoU} = \frac{|P \cap G|}{|P \cup G|},
\end{equation}
where $P$ and $G$ represent the predicted and ground-truth building regions, respectively.  
A higher IoU indicates better localization consistency.

\paragraph{Semantic Segmentation (SEG).}
For semantic segmentation, the mean Intersection over Union (mIoU) is used to evaluate the overall multi-class segmentation accuracy:
\begin{equation}
\mathrm{mIoU} = \frac{1}{C} \sum_{c=1}^{C} \frac{|P_c \cap G_c|}{|P_c \cup G_c|},
\end{equation}
where $C$ is the number of semantic classes, and $P_c$ and $G_c$ denote the predicted and ground-truth regions for class $c$.

\paragraph{Height Estimation (HE).}
For the height estimation task, the Root Mean Squared Error (RMSE) is employed to measure the deviation between the predicted and reference DSMs:
\begin{equation}
\mathrm{RMSE} = \sqrt{\frac{1}{N}\sum_{i=1}^{N}(H_i - \hat{H}_i)^2},
\end{equation}
where $H_i$ and $\hat{H}_i$ are the ground-truth and predicted height values for pixel $i$, respectively, and $N$ is the total number of valid pixels.

\section{Additional Experiments}
\label{app:addtion}

\subsection{Model Complexity}

\begin{table}[t]
\centering
\caption{Comparison of model complexity in terms of FLOPs and parameters.}
\label{tab:complexity}
\small
\renewcommand{\arraystretch}{1.1}
\setlength{\tabcolsep}{4pt}
\resizebox{\linewidth}{!}{
\begin{tabular}{l|cccccccc}
\toprule
\textbf{Model} & MPRNet & Restormer & AST & MambaIR & DFCFormer & EMRDM & GACR/2 & GACR/1 \\
\midrule
FLOPs (G)  & 548.65 & 140.99 & 38.86 & 206.80 & 138.45 & 166.72 & 56.05 & 223.60 \\
Params (M) & 3.64   & 26.10  & 19.90 & 3.16   & 29.98  & 39.13  & 73.81 & 73.80  \\
\bottomrule
\end{tabular}
}
\end{table}
As shown in Table~\ref{tab:complexity}, the GACR model with $p=2$ achieves a favorable trade-off between efficiency and capacity, requiring only 56.05G FLOPs while maintaining competitive performance among all compared methods. This compact configuration highlights the efficiency advantage of the proposed design. Although the $p=1$ variant can further improve the visual quality of cloud removal results, as reported in \cref{tab:quantative_results}, it introduces a substantial increase in computational cost (223.60G FLOPs). Therefore, we primarily adopt the $p=2$ configuration for comparison in subsequent experiments to maintain a balanced evaluation between effectiveness and efficiency.

\subsection{Visualization of t-SNE}

\begin{figure*}[htbp]
	\centering
	{\includegraphics[width=\linewidth]{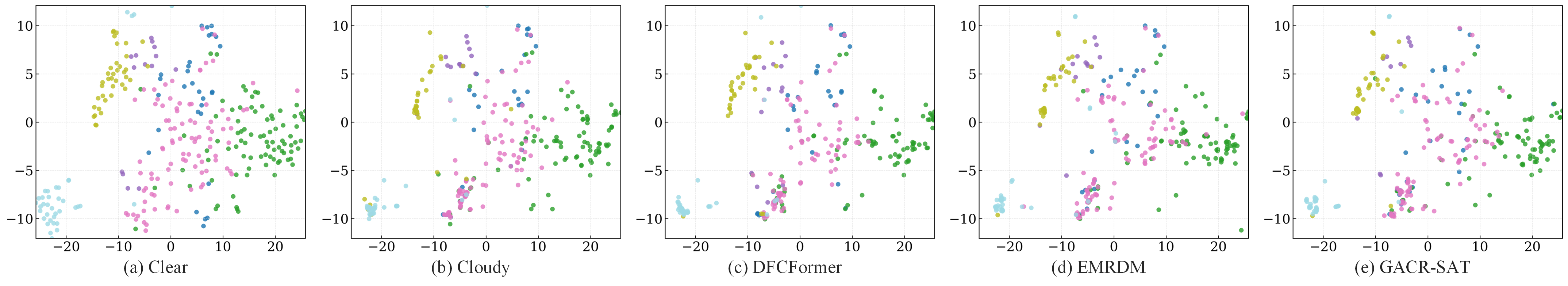}}
	\caption{The t-SNE visualization of feature representations on the CUHKCR-EXT-GZ dataset using the DINOv3 ViT-L/16-SAT-300M weights.}
	\label{fig:tsne}
\end{figure*}

We perform t-SNE visualization to analyze how CR influences the CLS task. Class tokens extracted from CUHKCR-EXT-GZ using the DINOv3 ViT-L/16-SAT-300M weights are projected into a t-SNE space fitted on clear-image features, as shown in \cref{fig:tsne}. The results show that even without CR, cloudy images already form well-separated clusters corresponding to different classes. Applying various CR methods does not further improve the separability between clusters. This phenomenon likely arises because classification primarily relies on global representations, which are only weakly affected by localized cloud noise. Consequently, the end-to-end approach can achieve performance comparable to, or even better than, most CR-based preprocessing pipelines.

\subsection{Downstream performance with DINOv3-SAT backbone}

\begin{table*}[t]
\centering
\caption{Quantitative comparison of downstream performance on ViT-L/16-SAT-300M weight across four downstream tasks, including classification (CLS), building extraction (BLD), semantic segmentation (SEG), and height estimation (HE).
The best and second-best results are highlighted in \textbf{bold} and \underline{underline}, respectively.}
\begin{adjustbox}{width=1.0\textwidth, center}
\begin{tabular}{l  |c c  |c c  |c llc  |c cll}
\toprule
\multirow{2}{*}{Model} & 
CLS-1&CLS-2&BLD-1&BLD-2& 
SEG-1&  SEG-2&SEG-3&SEG-4&HE-1& HE-2& HE-3&HE-4\\ 
& Acc. $\uparrow$& Acc. $\uparrow$& IoU $\uparrow$& IoU $\uparrow$& 
 mIoU $\uparrow$&   mIoU $\uparrow$&mIoU $\uparrow$&mIoU $\uparrow$& RMSE $\downarrow$&  RMSE $\downarrow$& RMSE $\downarrow$&RMSE $\downarrow$\\ 
\midrule
Upper Bound& 0.882 & 0.896 & 0.726 & 0.773 & 0.731 &   0.731 &0.742 &0.742 & 2.001 &  2.001 & 1.570 &1.570 
\\
 Lower Bound& 0.686 & 0.454 & 0.607 & 0.497 & 0.449 & 0.334 & 0.554 & 0.378 & 2.737 & 3.489 & 2.272 &2.949 
\\
 End-to-End& 0.830 & 0.890 & 0.694 & 0.709 & 0.686 & 0.593 & 0.706 & 0.623 & 2.217 & 2.622 & 1.715 &2.050 
\\
\midrule
 MPRNet& 0.498 & 0.500 & 0.654 & 0.534 & 0.677 & 0.559 & 0.675 & 0.569 & 2.280 & 2.905 & 1.711 &2.078 
\\
Restormer & 0.542 & 0.533 & 0.664 & 0.572 & 0.690 &   0.591 &0.677 &0.600 & 2.166 &  2.540 & 1.628 &1.889 
\\
AST & 0.415 & 0.435 & 0.673 & 0.620 & 0.636 &   0.469 &0.629 &0.490 & 2.449 &  3.143 & 1.799 &2.406 
\\
MambaIR & 0.458 & 0.484 & 0.667 & 0.501 & 0.682 &   0.547 &0.659 &0.602 & 2.241 &  2.910 & 1.688 &1.921 
\\
DFCFormer & 0.639 & 0.500 & 0.667 & 0.555 & 0.681 &   0.574 &0.663 &0.621 & 2.225 &  2.679 & 1.656 &1.876 
\\
EMRDM & 0.642 & 0.647 & \underline{0.699}& 0.681 & 0.709 &   0.641 &\underline{0.730}&0.634 & 2.092 &  2.353 & 1.599 &1.830 
\\
\midrule
GACR-SAT& \underline{0.753}& \underline{0.729}& \textbf{0.700}& \textbf{0.707}& \textbf{0.727}&   \textbf{0.676}&\textbf{0.740}&\underline{0.692}& \textbf{2.061}&  \underline{2.236}& \textbf{1.582}&\textbf{1.682}\\
GACR-LVD& \textbf{0.833}& \textbf{0.742}& 0.697 & \underline{0.702}& \underline{0.715}&   \underline{0.673}&0.727 &\textbf{0.698}& \underline{2.062}&  \textbf{2.229}& \underline{1.589}&\underline{1.682}\\
\bottomrule
\end{tabular}
\end{adjustbox}
\label{tab:dst_compare_sat}
\end{table*}

In \cref{tab:dst_compare}, we report the results obtained using ViT-L/16-LVD-1689M as the pretrained backbone for downstream networks. We further evaluate the performance using the ViT-L/16-SAT-300M weights. As shown in \cref{tab:dst_compare_sat}, GACR consistently outperforms other methods under this downstream setting, further demonstrating the effectiveness of the proposed framework. Notably, we observe that most models exhibit a certain degree of performance degradation when switching from LVD to SAT weights, which may be attributed to the richer and more diverse pretraining data used by LVD.

These results provide a practical insight into the use of different DINOv3 weights for remote sensing tasks. The SAT weights, pretrained on remote sensing datasets, capture more domain-specific representations that benefit low-level tasks such as image restoration. In contrast, the LVD weights, pretrained on a broader and more diverse corpus, offer stronger semantic understanding capabilities and are therefore more suitable for high-level downstream tasks. Additionally, differences in the input resolution used during pretraining may also contribute to the observed domain discrepancy between the two pretrained models.

\subsection{Evaluation on Heterogeneous Downstream Architectures.}
To further examine the robustness of the proposed framework with respect to downstream model architectures, we conduct additional experiments using two heterogeneous segmentation networks: the ResNet-based $A^{2}$-FPN and the CNN–Transformer hybrid UNetFormer. Both networks are trained on cloud-free data following the same evaluation protocol described in the main text. The quantitative results on the Vaihingen-CR dataset are summarized in Table~R1.

As shown in \cref{tab:a2fpn}, GACR consistently achieves the best or second-best performance across all configurations. On the Vaihingen-CR-thin setting, GACR-SAT obtains mIoU scores of 0.739 and 0.678 when evaluated with $A^{2}$-FPN and UNetFormer, respectively. Similar improvements are observed under the more challenging Vaihingen-CR-thick condition, where GACR-SAT reaches 0.721 and 0.662. These results demonstrate that the performance gains brought by GACR are not tied to a specific downstream backbone but remain effective across heterogeneous architectures.

This behavior suggests that the proposed geo-contextual alignment helps restore semantically meaningful structures that are beneficial for recognition, enabling consistent improvements across different downstream models.

\begin{table}[t]
\centering
\caption{Quantitative comparison of SEG tasks with $A^{2}$-FPN (ResNet-based) and UNetFormer (CNN-Transformer Hybrid).}
\label{tab:a2fpn}
\resizebox{0.7\textwidth}{!}{
\begin{tabular}{l|cc|cc}
    \toprule
    \multirow{2}{*}{Method} &
    \multicolumn{2}{c|}{Vaihingen-CR-thin} &
    \multicolumn{2}{c}{Vaihingen-CR-thick} \\
    \cmidrule(lr){2-3}\cmidrule(lr){4-5}
    & $A^{2}$-FPN & UNetFormer & $A^{2}$-FPN & UNetFormer \\
    \midrule
    Upper Bound        & 0.747 & 0.687 & 0.747 & 0.687 \\
    Lower Bound        & 0.384 & 0.289 & 0.277 & 0.231 \\
    \midrule
    MPRNet             & 0.713 & 0.623 & 0.619 & 0.527 \\
    Restormer          & 0.731 & 0.653 & 0.687 & 0.599 \\
    AST                & 0.718 & 0.622 & 0.587 & 0.513 \\
    MambaIR            & 0.725 & 0.642 & 0.668 & 0.566 \\
    DFCFormer          & 0.728 & 0.645 & 0.695 & 0.617 \\
    EMRDM              & 0.731 & 0.655 & 0.688 & 0.584 \\
    \midrule
    GACR-SAT/2    & \textbf{0.739} & \textbf{0.678} & \textbf{0.721} & \textbf{0.662} \\
    GACR-LVD/2    & \underline{0.739} & \underline{0.677} & \underline{0.719} & \underline{0.653} \\
    \bottomrule
\end{tabular}
}
\end{table}

\subsection{CKNNA with Cloud-free Data}

To provide a more intuitive comparison of the representational consistency between CR images and their cloud-free counterparts, we compute the CKNNA~\cite{huh2024prh} scores between each CR result and its corresponding cloud-free reference for all competing methods. As shown in Table~\ref{tab:cknna}, the CKNNA values of most CR results are generally higher than those of the original cloudy images, indicating improved feature-level alignment after cloud removal. Notably, on the Potsdam-CR-thick dataset, some methods yield slightly lower CKNNA scores than the cloudy images, suggesting that their reconstructed results deviate more from the cloud-free representations. In contrast, GACR achieves consistently higher CKNNA scores across most datasets, demonstrating its stronger capability to enhance representation consistency between the restored and cloud-free images.

\begin{table}[t]
\centering
\caption{Comparison of CKNNA between CR images and cloud-free references across six datasets.}
\label{tab:cknna}
\small
\renewcommand{\arraystretch}{1.12}
\setlength{\tabcolsep}{4pt}
\resizebox{\columnwidth}{!}{
\begin{tabular}{l|cccccc}
\toprule
\textbf{Method} & CUHKCR-EXT-GZ & CUHKCR-EXT-CS & Potsdam-CR-thin & Potsdam-CR-thick & Vaihingen-CR-thin & Vaihingen-CR-thick \\
\midrule
cloudy            & 0.319 & 0.330 & 0.472 & 0.584 & 0.592 & 0.477 
\\
MPRNet            & 0.328 & 0.362 & 0.641 & 0.535 & 0.661 & 0.566 
\\
Restormer         & 0.338 & 0.371 & 0.666 & 0.564 & 0.699 & 0.596 
\\
AST               & 0.334 & 0.363 & 0.618 & 0.506 & 0.628 & 0.521 
\\
MambaIR           & 0.328 & 0.362 & 0.654 & 0.538 & 0.679 & 0.588 
\\
DFCFormer         & 0.338 & 0.373 & 0.657 & 0.553 & 0.689 & 0.598 
\\
EMRDM             & 0.346 & 0.379 & 0.700 & 0.592 & 0.727 & 0.613 
\\
\midrule
GACR-SAT& \textbf{0.352} & \textbf{0.388} & \textbf{0.710} & \textbf{0.616} & \textbf{0.747} & \textbf{0.665} 
\\
GACR-LVD& \underline{0.347} & \underline{0.386} & \underline{0.706} & \underline{0.610} & \underline{0.739} & \underline{0.657} 
\\
\bottomrule
\end{tabular}} 
\end{table}

\subsection{Additional Visual Results}
\begin{figure*}[htbp]
	\centering
	{\includegraphics[width=0.9\linewidth]{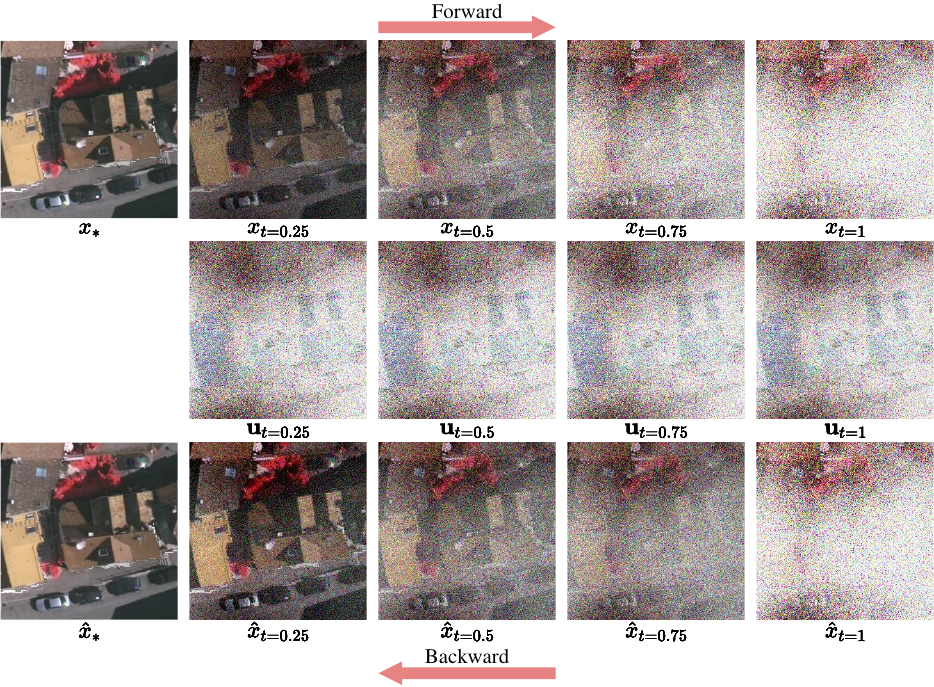}}
	\caption{Visualization of the forward and reverse processes of the OAR-Flow model.}
	\label{fig:viz_step}
\end{figure*}

\begin{figure*}[htbp]
	\centering
	{\includegraphics[width=0.9\linewidth]{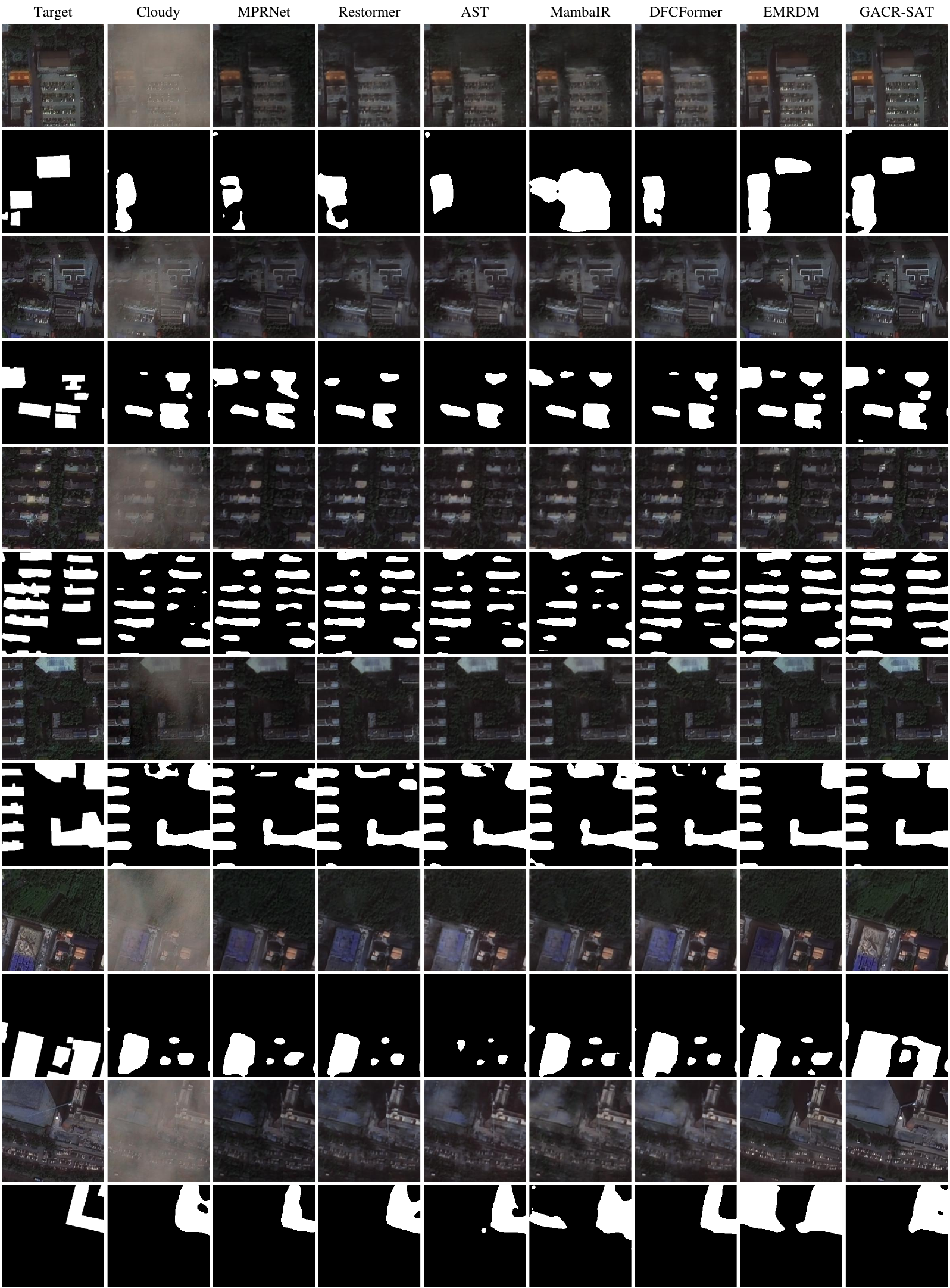}}
	\caption{Additional CR results on CUHKCR-EXT-GZ and the corresponding BLD results.}
	\label{fig:add_viz_gz}
\end{figure*}

\begin{figure*}[htbp]
	\centering
	{\includegraphics[width=0.9\linewidth]{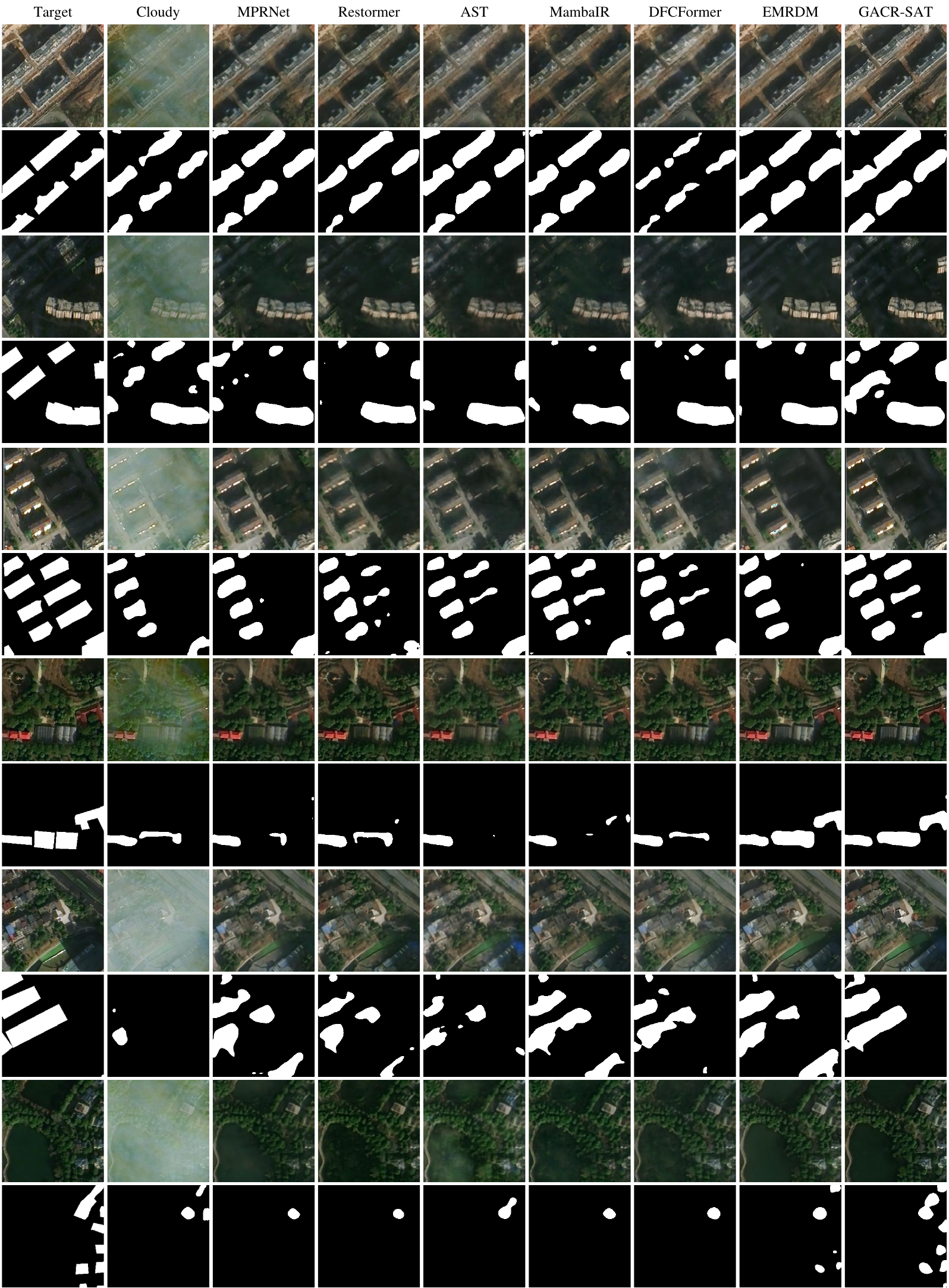}}
	\caption{Additional CR results on CUHKCR-EXT-CS and the corresponding BLD results.}
	\label{fig:add_viz_cs}
\end{figure*}

\begin{figure*}[htbp]
	\centering
	{\includegraphics[width=0.9\linewidth]{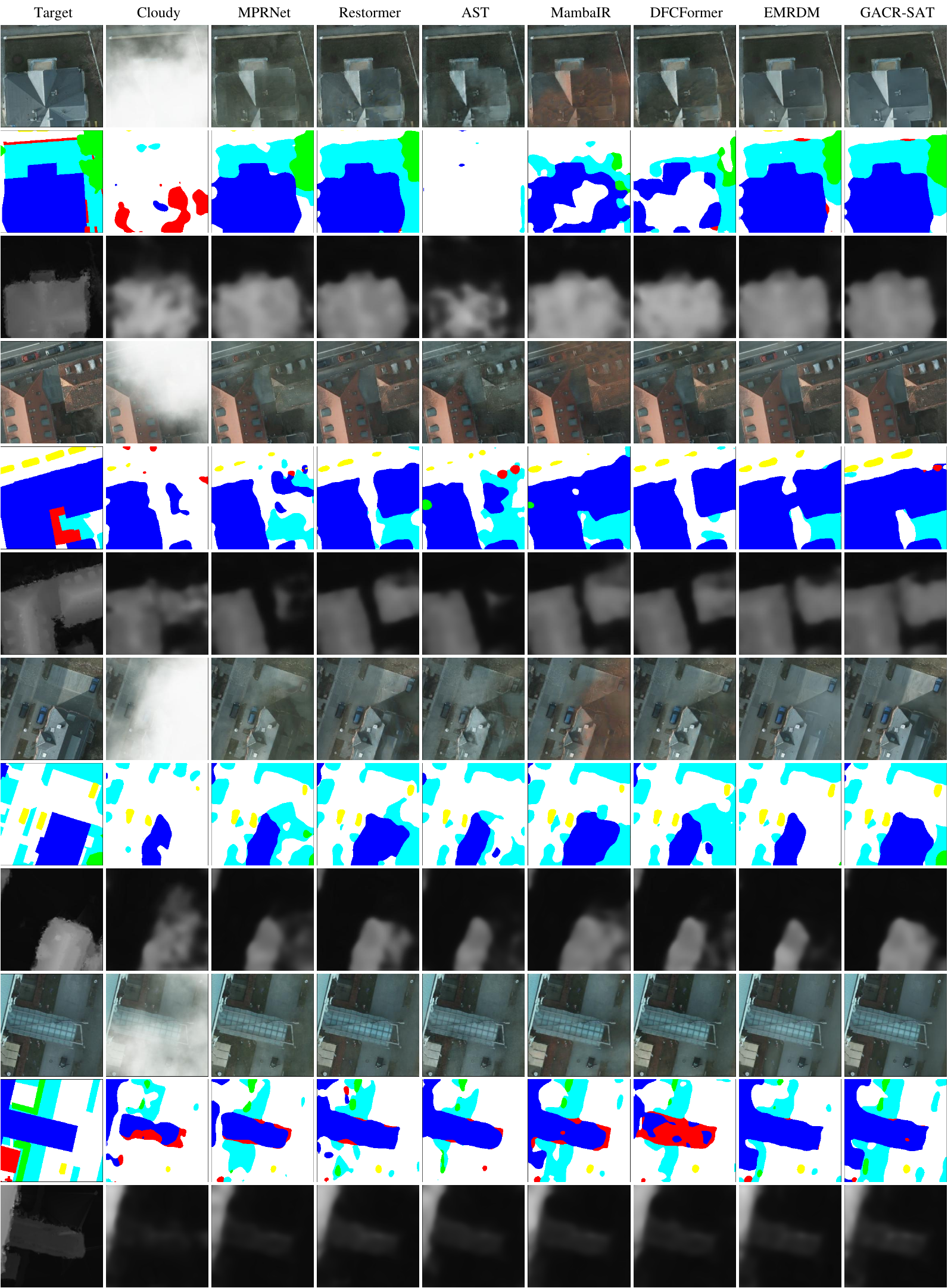}}
	\caption{Additional CR results on Potsdam-CR-thick and the corresponding SEG and HE results.}
	\label{fig:add_viz_potsdam}
\end{figure*}

\begin{figure*}[htbp]
	\centering
	{\includegraphics[width=0.9\linewidth]{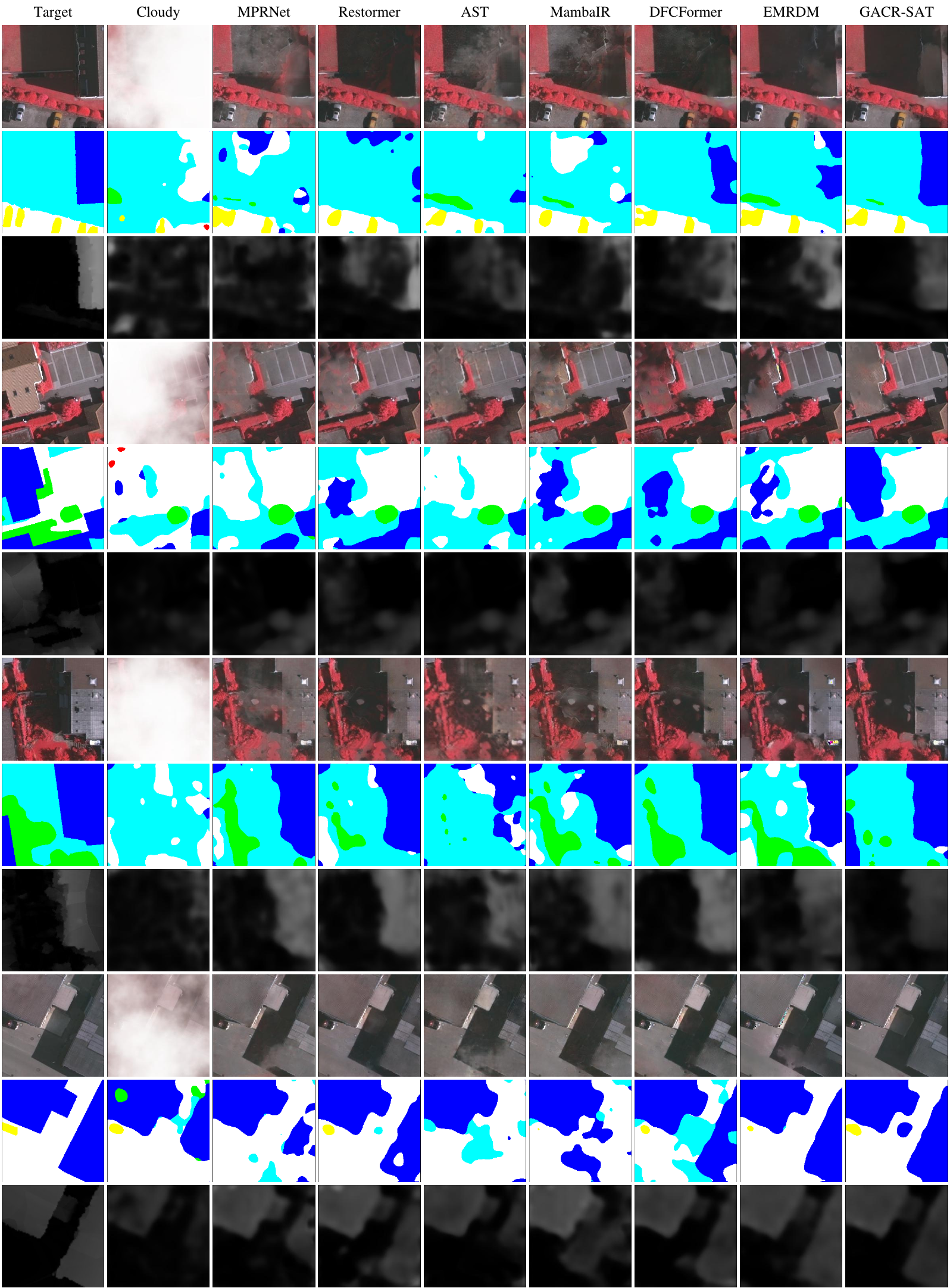}}
	\caption{Additional CR results on Vaihingen-CR-thick and the corresponding SEG and HE results.}
	\label{fig:add_viz_vaihingen}
\end{figure*}

In this section, we present additional visualization results. \cref{fig:viz_step} illustrates the detailed forward and reverse processes of OAR-Flow. \cref{fig:add_viz_gz}, \cref{fig:add_viz_cs}, \cref{fig:add_viz_potsdam}, and \cref{fig:add_viz_vaihingen} provide additional visual comparisons on the CUHKCR-EXT-GZ, CUHKCR-EXT-CS, Potsdam-CR-thick, and Vaihingen-CR-thick datasets, respectively.

\subsection{Additional Ablation Studies}

\textbf{Effect of VFM backbones.}
To examine whether the effectiveness of GCPA depends on a specific visual foundation model, we replace the DINOv3 encoder with DINOv2, CLIP, and MAE while keeping OAR-Flow unchanged. As shown in Tab.~\ref{tab:vfm_backbone_ablation}, all VFM variants consistently outperform the baseline, indicating that the improvement is not tied to a particular encoder choice. DINOv3 achieves the best overall performance, while DINOv2 and CLIP also provide clear gains, confirming the generality of using semantic priors for cloud removal.

\begin{table}[t]
\centering
\caption{Ablation study of different VFM backbones on Vaihingen-CR-thick.}
\label{tab:vfm_backbone_ablation}
\begin{tabular}{l|cccc}
\toprule
Backbone & PSNR$\uparrow$ & SSIM$\uparrow$ & SEG$\uparrow$ & HE$\downarrow$ \\
\midrule
DINOv2 & 32.552 & 0.962 & 0.726 & 1.564 \\
CLIP   & 32.203 & 0.962 & 0.725 & 1.566 \\
MAE    & 31.096 & 0.955 & 0.720 & 1.606 \\
\midrule
Baseline & 28.979 & 0.951 & 0.692 & 1.629 \\
\bottomrule
\end{tabular}
\end{table}

\textbf{Effect of the anchoring strength $\rho$.}
We further study the influence of the observation anchoring strength $\rho$ in OAR-Flow. As shown in Tab.~\ref{tab:rho_ablation}, increasing $\rho$ from 0 to 3 improves the reconstruction quality, with PSNR increasing from 32.264 to 33.018. This suggests that moderate anchoring helps preserve reliable observation cues and stabilizes the reverse trajectory. When $\rho$ is further increased, the performance slightly drops, indicating that overly strong anchoring may restrict the model's ability to recover heavily obscured regions. We therefore adopt $\rho=3$ by default.

\begin{table}[t]
\centering
\caption{Ablation study of the anchoring strength $\rho$ on Vaihingen-CR-thick.}
\label{tab:rho_ablation}
\setlength{\tabcolsep}{8pt}
\begin{tabular}{c|cccccc}
\toprule
$\rho$ & 0 & 1 & 2 & 3 & 4 & 5 \\
\midrule
PSNR$\uparrow$ & 32.264 & 32.648 & 33.006 & 33.018 & 32.876 & 32.672 \\
\bottomrule
\end{tabular}
\end{table}

\end{document}